\documentclass{mosi}

\usepackage{pifont}      
\usepackage{enumitem}    
\usepackage{float}       
\usepackage{makecell}    
\usepackage{tabularx}    
\usepackage{algorithm}        
\usepackage{algpseudocode}    
\crefname{algorithm}{Algorithm}{Algorithms}
\Crefname{algorithm}{Algorithm}{Algorithms}

\graphicspath{{content/}{assets/}{./}}
\makeatletter
\def\input@path{{content/}}
\makeatother

\definecolor{oursgray}{gray}{0.95}   
\definecolor{tickG}{HTML}{00C853}
\definecolor{crossR}{HTML}{FF1744}

\newtcolorbox{promptbox}[2][]{
  colback=white, coltext=black,
  arc=3mm, boxrule=0.5pt, colframe=black!60!white,
  title={#2}, colbacktitle=black, coltitle=white, fonttitle=\bfseries,
  top=8pt, bottom=8pt, left=10pt, right=10pt,
  breakable,
  before upper={\linespread{1}\selectfont
    \setlength{\parskip}{1ex plus 0.2ex minus 0.2ex}\setlength{\parindent}{0pt}},
  #1
}

\usepackage{bm,xspace,fontawesome5}

\usepackage{amsmath,amsfonts,bm}

\def\eqref#1{equation~\ref{#1}}

\def\1{\bm{1}}

\DeclareMathAlphabet{\mathsfit}{\encodingdefault}{\sfdefault}{m}{sl}
\SetMathAlphabet{\mathsfit}{bold}{\encodingdefault}{\sfdefault}{bx}{n}

\newcommand{\ours}{\textbf{MOSS-Image}\xspace}
\newcommand{\llb}{\textbf{LLTB}\xspace}
\newcommand{\rla}{\textbf{RLA}\xspace}

\colorlet{insightaccent}{mosiblue}
\colorlet{insightbg}{mosicyan!8!white}
\newtcolorbox{insightbox}{%
  enhanced, breakable, colback=insightbg, frame hidden,
  sharp corners, boxrule=0pt,
  borderline west={2.2pt}{0pt}{insightaccent},
  left=3.2mm, right=3mm, top=1.6mm, bottom=1.6mm,
  before skip=0.8\baselineskip, after skip=0.8\baselineskip,
  fontupper=\normalfont\normalsize,
}
\newcommand{\insight}[1]{%
  \begin{insightbox}
    \hangindent=1.4em\hangafter=1%
    \noindent\makebox[1.4em][l]{\textcolor{insightaccent}{\faLightbulb[regular]}}%
    #1\par
  \end{insightbox}%
}

\newcommand{\hficon}{%
  \raisebox{-0.15em}{%
    \includegraphics[height=1em]{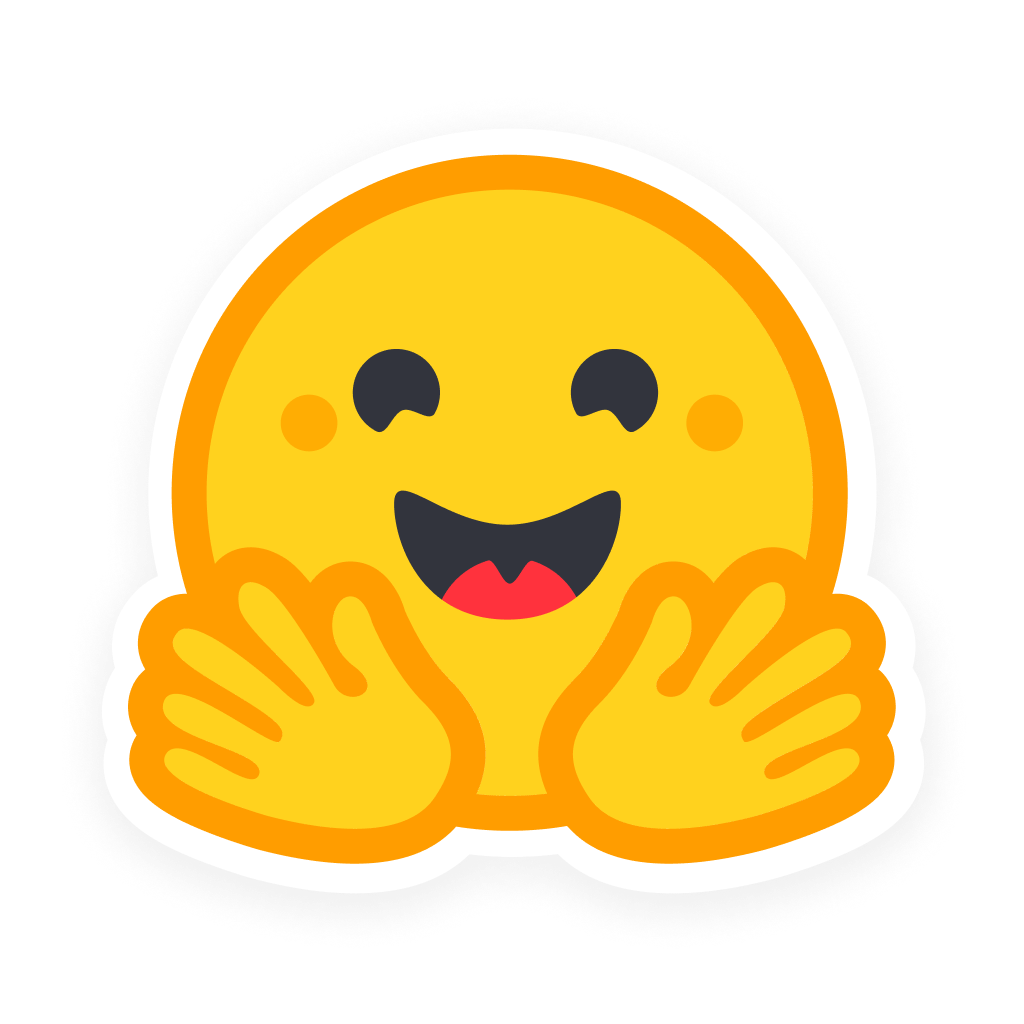}}}

\title{MOSS-Image Technical Report}

\author[1,2,3,4,*]{MOSI.AI}

\affil[1]{MOSI Intelligence}
\affil[2]{SII}
\affil[3]{Fudan University}
\affil[4]{Shanghai Jiao Tong University}

\abstract{
Recent closed-source image generation systems can produce diverse visual content from simple compositions to information-dense graphics, yet open-weight text-to-image models still struggle with dense text and intricate layouts.
To narrow this gap, we introduce \ours, an open generation system that combines high-fidelity visual synthesis with intent-aware composition.
First, we pretrain a 5B single-stream diffusion transformer (DiT) from scratch with a carefully designed recipe, enabling bilingual long-form text rendering and precise layout control.
Second, we develop an intention-aware agentic prompt enhancement (PE) module that translates diverse user requests into detailed content and layout specifications.
Together, they enable \ours to generate text-rich and layout-coherent images.
We further introduce LongLongTextBench (\llb) for evaluating long-form English and Chinese text rendering, and RealLayoutArena (\rla) for end-to-end layout-rich generation. 
\ours achieves state-of-the-art performance among open-weight models on \llb, with an average text coverage of 96.3\%. 
On \rla, \ours ranks second in the end-to-end system comparison, while its backbone ranks first among generators evaluated with shared enhanced prompts.
We will release the model, training recipe, and evaluation framework to support further research and promote the text-to-image community.
}

\checkdata[\faGithub\enspace GitHub]{\url{https://github.com/OpenMOSS/MOSS-Image}}
\checkdata[\faGlobe\enspace Online Demo]{\url{https://huggingface.co/spaces/OpenMOSS-Team/MOSS-Image}}
\checkdata[\hficon\enspace Hugging Face]{\url{https://huggingface.co/collections/OpenMOSS-Team/moss-image}}

\begin{document}
\maketitle
\authornote{$^{*}$Full contributors can be found in the Contributors section.}

\section{Introduction}
\label{sec:introduction}

\begin{figure*}[t]
\centering
\includegraphics[width=\linewidth]{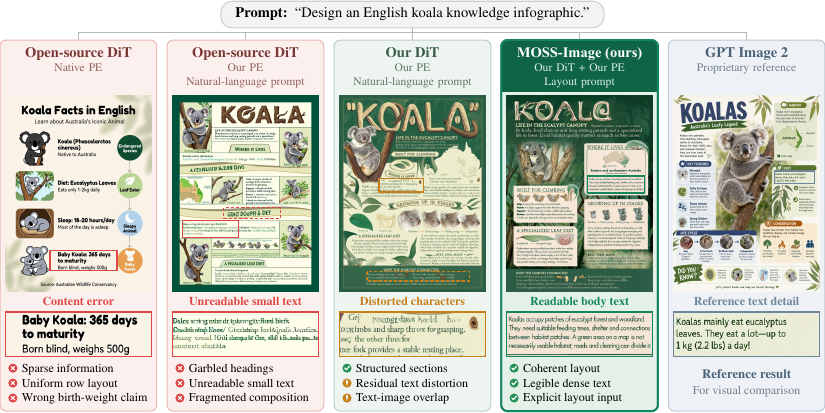}
\caption{\textbf{Complex graphic generation from a concise user request.} The examples illustrate differences in content, composition, and text rendering across generation configurations. \ours combines a pretrained diffusion transformer with prompt enhancement and optional explicit layout specifications. The proprietary output provides a visual reference; quantitative comparisons are reported in Section~\ref{sec:experiments}.}
\label{fig:motivation}
\end{figure*}

Text-to-image (T2I) foundation models have made substantial progress in visual quality, semantic alignment, and text rendering in recent years.
Proprietary systems, such as GPT Image~2~\citep{openai_gpt_image_2}, Seedream~5.0 Pro~\citep{seedream_5_pro}, Qwen Image~3.0 Pro~\citep{qwen_image_3}, and Nano Banana Pro~\citep{google_nano_banana_pro} have extended image generation from artistic creation to practical visual communication, such as infographics, user interfaces, and knowledge cards. 
Open-weight models, including Stable Diffusion~\citep{rombach2021highresolution}, SDXL~\citep{podell2024sdxl}, FLUX~\citep{flux-2-2025}, Qwen-Image~\citep{wu2025qwen}, Z-Image~\citep{cai2025z}, ERNIE-Image~\citep{liu2026ernie}, and Boogu-Image~\citep{chen2026boogu}, have made these capabilities increasingly accessible, yet they still struggle with dense text and intricate layouts.
For such graphics, a plausible overall appearance does not guarantee legible text, complete content, or coherent spatial organization (Figure~\ref{fig:motivation}). 
At the same time, a concise request may specify only a topic and format, leaving the actual content, information hierarchy, and spatial organization to be developed.

We attribute this gap to two coupled challenges: \textit{rendering} and \textit{specification}. \textit{Rendering} involves accurately reproducing long text and following spatial constraints. \textit{Specification} involves translating a concise user request into content and layout instructions, with a level of detail appropriate to the request.
Neither capability alone is sufficient: detailed specifications do not guarantee faithful rendering, while rendering ability alone does not resolve the content and design decisions left unspecified by the user.
This calls for a system that connects intent-aware specification with faithful rendering, and for evaluation that assesses the resulting designs end to end.



To address these issues, we present \ours, which pairs a text-and-layout-capable generator with an agent that enriches sparse requests and designs deliberate layouts. 
On the generation side, we pretrain a 5B single-stream diffusion transformer (DiT) from scratch for dense bilingual text and multi-element graphics. Its recipe combines broad data retention, multi-granularity and structure-aware textual supervision, and progressive-resolution training; crucially, natural-language descriptions and explicit layout specifications are learned through a single text-conditioning pathway, so the model can follow either form. On the composition side, we develop an intention-aware agentic prompt enhancement (PE) module that routes each request to task-specific guidance and retrieves knowledge and visual references as needed. Unlike such generator-agnostic agents~\citep{zhang2026qwen}, which can only communicate through natural-language prompts, our enhancer turns brief requests into structured layouts that the DiT follows for precise composition control.



To evaluate rendering, we introduce LongLongTextBench (\llb) for long-form English and Chinese text, complementing LongTextBench and 7Bench. These benchmarks assess the execution of specified text or spatial constraints, but do not measure whether a complete system can select and organize content into a coherent design from a concise user request. We therefore introduce RealLayoutArena (\rla) to evaluate end-to-end generation of complex graphics from short design requests. Since a request can admit multiple valid designs, \rla uses blind pairwise human comparisons to jointly assess instruction following, content completeness, text readability, and visual organization. We further conduct prompt-strategy ablations and cross-backbone comparisons under shared enhanced prompts to disentangle the roles of specification and generator. 


On \llb, \ours achieves state-of-the-art performance among open-weight models with a Mean-6 score of 96.3, surpassing Qwen Image~2512 (64.9) by 31.4 percentage points. On \rla, \ours achieves 1698.70 Elo, ranking second in the end-to-end comparison, behind GPT Image~2 and ahead of Seedream~5.0 Pro. With shared enhanced natural-language prompts, our backbone achieves the highest rating at 1807.34 Elo. Prompt comparisons further show that detailed specifications substantially improve human preference over concise requests.

Our contributions are summarized as follows:
\begin{itemize}
\item \textbf{A pretrained foundation model for information-dense graphics.} We develop a 5B diffusion transformer pretrained from scratch that demonstrates strong bilingual long-form text rendering and layout control through a unified text-conditioning interface. We detail its pretraining recipe, including broad data retention, multi-granularity and structure-aware textual supervision, and progressive-resolution training, providing a practical foundation for further research on complex graphic generation.

\item \textbf{An agentic end-to-end generation system.} We build \ours by pairing the foundation model with an intention-aware agentic prompt enhancer that translates diverse user requests, even those of fewer than ten words, into detailed, executable generation specifications. Through task-specific routing, knowledge and visual reference retrieval, and structured layout planning, the system produces graphics that remain faithful to user intent while adapting their information density to what each request calls for, from minimalist designs to information-rich layouts.

\item \textbf{Benchmarks and systematic analysis.} We introduce \llb, which evaluates bilingual text rendering under extended text lengths, and \rla, which evaluates complex graphic generation under realistic user requests at the system level. \ours achieves state-of-the-art open-weight performance on \llb (96.3 Mean-6), ranks second in the end-to-end \rla comparison, and attains the highest rating under shared enhanced prompts. Prompt comparisons further assess the effects of content and layout details.

\end{itemize}

\section{System Overview}
\label{sec:model}

\begin{figure*}[t]
\centering
\includegraphics[width=\linewidth]{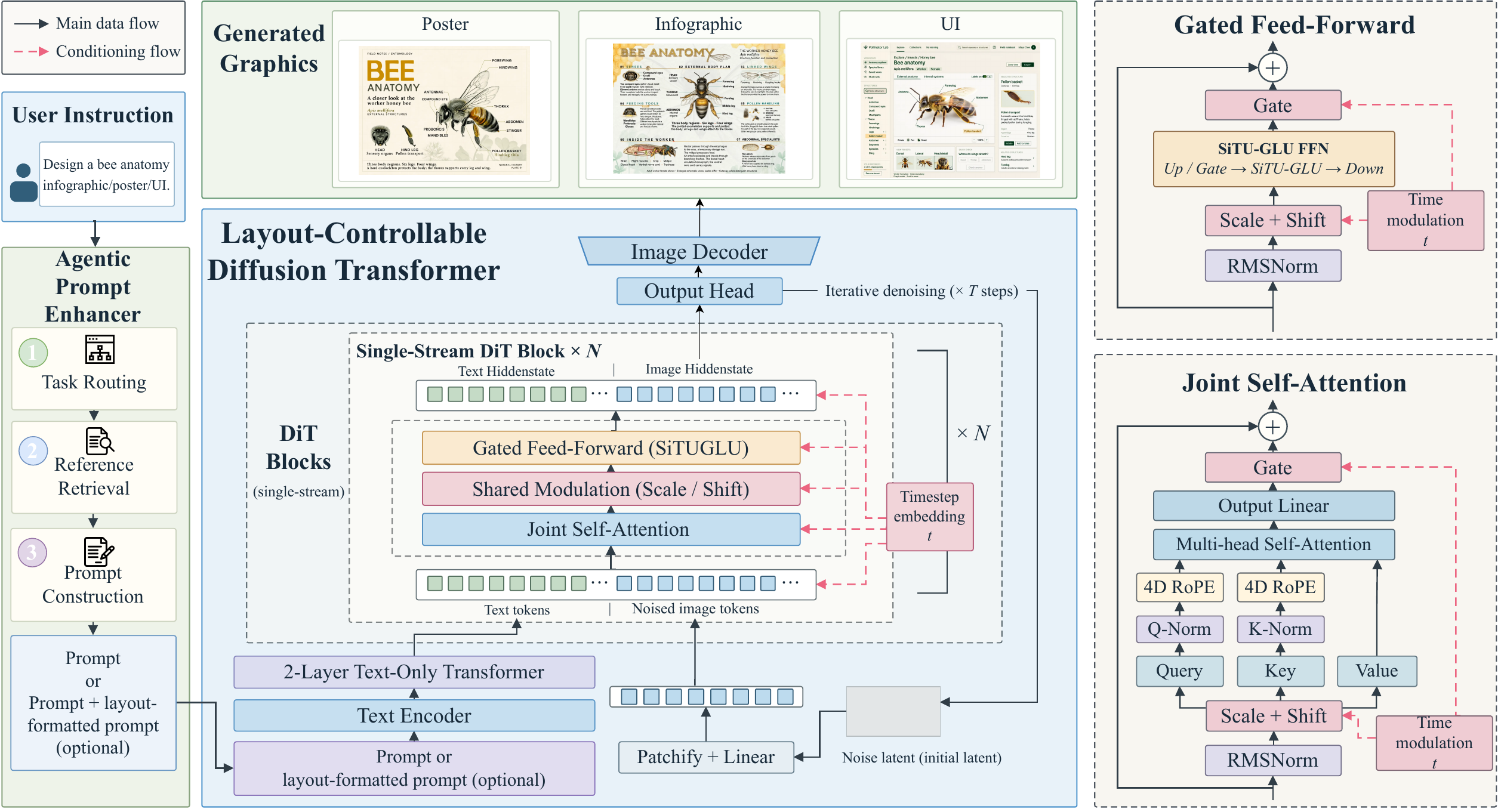}
\caption{\textbf{Overview of \ours.} An intent-aware prompt enhancer constructs content and layout specifications from user requests, and a pretrained 5B single-stream DiT renders them into images through a shared text-conditioning interface. The right panel details the renderer's joint self-attention and gated feed-forward components.}
\label{fig:overview}
\end{figure*}

\ours pairs a pretrained text-and-layout generator with an intent-aware prompt enhancer (Figure~\ref{fig:overview}). The enhancer turns a concise request into a content and layout specification; the generator renders that specification into an image.

\paragraph{Text and layout rendering.}
\label{sec:rendering}
\label{sec:conditioning}
To render long text and follow spatial constraints, we pretrain a 5B single-stream DiT from scratch. Natural-language descriptions and structured layouts share one text-conditioning pathway. Broad data retention, multi-granularity and structure-aware supervision, and progressive training build this capability. Section~\ref{sec:training} details the model pretraining recipe.

\paragraph{Intent-aware prompt enhancement.}
To resolve the content and design choices left unspecified by a short request, an agent selects task-specific guidance, retrieves knowledge and visual references as needed, and constructs detailed generation specifications. Its layout output uses the same textual interface learned by the generator. Section~\ref{sec:pe} describes this inference-time module.

\section{Pretraining Strategy}
\label{sec:training}

\begin{figure*}[t]
\centering
\includegraphics[width=\linewidth]{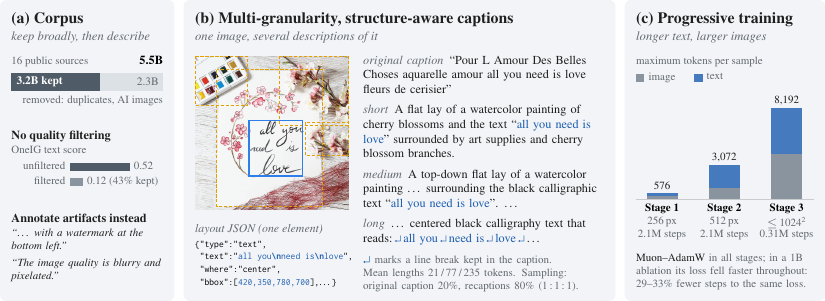}
\caption{\textbf{Pretraining recipe of \ours.} (a) After deduplication and AI-image removal, we keep all images and caption their artifacts instead of filtering them out. (b) A ReLAION sample with its original caption, three recaptions, and layout JSON (excerpts); the JSON's object and text boxes are drawn in orange and blue. Visible text, marked in blue, is transcribed verbatim, line breaks ({\usefont{U}{psy}{m}{n}\char191}) included. (c) Maximum tokens per sample in each stage.}
\label{fig:training}
\end{figure*}

We pretrain a general-purpose text-to-image foundation model and evaluate the resulting backbone on long-form text rendering and layout control. Figure~\ref{fig:training} summarizes the recipe: corpus construction, multi-granularity and structure-aware textual supervision, and multi-stage DiT pretraining. This recipe develops the rendering module in Section~\ref{sec:rendering}; at inference, the enhancer in Section~\ref{sec:pe} supplies content and layout specifications through the same conditioning interface.

\paragraph{Backbone architecture.}
\label{sec:architecture}
The 5B single-stream DiT jointly processes text tokens from a frozen Qwen3.5-9B encoder and image latents from a frozen FLUX.2 VAE. Its blocks use joint self-attention, shared modulation~\citep{flux2024}, and SiTU-GLU feed-forward layers~\citep{team2026kimi}. For an input $\mathbf{x}$, let $\mathbf{g}=W_g\mathbf{x}$ and $\mathbf{u}=W_u\mathbf{x}$. With the smooth cap $c_\beta(v)=\beta\tanh(v/\beta)$, the SiTU-GLU layer computes
\begin{equation}
\label{eq:situ_glu}
\operatorname{FFN}(\mathbf{x})=W_o\!\left[c_{\beta_g}(\mathbf{g})\odot\operatorname{sigmoid}(\mathbf{g})\odot c_{\beta_u}(\mathbf{u})\right],
\end{equation}
where $W_g,W_u,W_o$ are learned projections, $\beta_g,\beta_u>0$ control saturation, and $\odot$ denotes elementwise multiplication.

For 4D RoPE, each query or key head is partitioned as $\mathbf{z}=[\mathbf{z}_r;\mathbf{z}_t;\mathbf{z}_w;\mathbf{z}_h]$ and assigned coordinates $\mathbf{p}=(r,t,w,h)$ for reference index, sequence position, width, and height. We apply independent rotations to the four subspaces:
\begin{equation}
\label{eq:rope_4d}
\begin{aligned}
\operatorname{RoPE}_{4D}(\mathbf{z},\mathbf{p})
&=[R_r(r)\mathbf{z}_r;R_t(t)\mathbf{z}_t;R_w(w)\mathbf{z}_w;R_h(h)\mathbf{z}_h],\\
R_a(p_a)&=\operatorname{diag}_{j}\!\begin{pmatrix}
\cos(p_a\omega_{a,j})&-\sin(p_a\omega_{a,j})\\
\sin(p_a\omega_{a,j})&\cos(p_a\omega_{a,j})
\end{pmatrix}.
\end{aligned}
\end{equation}
Here $a\in\{r,t,w,h\}$, $j$ indexes feature pairs, $\omega_{a,j}$ is the corresponding rotary frequency, and $\operatorname{diag}_{j}$ assembles the rotation blocks. The coordinate $t$ is a positional index, distinct from the diffusion timestep. Without reference inputs, $r=0$ gives $R_r(0)=I$, leaving the reference subspace unrotated and yielding partial RoPE. Appendix~\ref{app:model_configuration} gives the model configuration.

\subsection{Pretraining Corpus Construction}
\label{sec:data}

We construct the corpus from ReLAION~\citep{relaion}, OmniCorpus~\citep{li2024omnicorpus}, DataComp~\citep{gadre2023datacomp}, and additional public image-text datasets. Following source-specific build preprocessing, corpus preparation uses SSCD-based deduplication~\citep{pizzi2022self} and removal of AI-generated images using an in-house detector, reducing the corpus from approximately 5 billion to 3 billion images. We mix data sources in their original proportions.

\paragraph{No additional data filtering.}
At this subsequent corpus-preparation stage, we apply no further filtering beyond deduplication and AI-image removal; this does not exclude quality filters already applied during source builds, such as the ReLAION NSFW and watermark checks. This is a deliberate pretraining choice, supported by preliminary filtering experiments, to retain broad visual coverage. Residual attributes such as watermarks, blurry faces, and compression artifacts are incorporated into the textual supervision rather than used as exclusion criteria. Appendix~\ref{app:data_cons} provides dataset details, and Appendix~\ref{app:filtering_study} describes the preliminary experiments.

\insight{\textbf{Annotate rather than filter.} We advocate preserving real-world data distributions by annotating undesirable attributes rather than filtering out images.}

\subsection{Multi-granularity and Structure-aware Supervision}
\label{sec:supervision}

\paragraph{Descriptions at multiple granularities.}
We caption images with gemma4~\citep{team2026gemma} at different semantic granularities: a short sentence, a medium paragraph, or a detailed description of composition, subjects, scene attributes, and visible text. These variants are defined by content rather than fixed token budgets. Training samples among available descriptions using domain-specific policies. Appendix~\ref{app:caption_sampling} details generation recipes, length statistics, and conditional sampling.

\paragraph{Explicit text and structure.}
For a subset of images, we additionally construct structured annotations containing global descriptions and localized elements with bounding boxes, categories, and, where applicable, transcribed text and typography attributes. The documented annotation recipes validate JSON structure and coordinate constraints. During training, annotations are geometrically aligned and optional fields undergo stochastic dropout before serialization through the text-conditioning interface in Section~\ref{sec:conditioning}. Appendix~\ref{app:caption_validation} describes annotation checks, and Appendix~\ref{app:conditioning} details the schemas and preprocessing policies.

\paragraph{Preserving line structure.}
We preserve line breaks in annotated text to match its arrangement in the image. This is important because accurate transcription specifies which words appear, while line breaks additionally encode how they are organized across lines. Aligning the annotation with the visible line structure provides explicit supervision for multi-line text rendering. Section~\ref{sec:linebreak_analysis} examines this design through a case study.

\insight{\textbf{Text line breaks are also a form of layout.}}

\subsection{Pretraining Schedule and Optimization}
\label{sec:training_schedule}


\paragraph{Progressive training.}
Training increases image resolution and text-context capacity across three stages (Table~\ref{tab:training_schedule}). The final stage preserves aspect ratio, aligns image dimensions to multiples of 16, and caps target image area at $1024^2$ pixels. We concatenate the condition fields, apply the Qwen chat template, and tokenize with a final-stage limit of 4096 text tokens.

\begin{table}[t]
\centering
\small
\setlength{\tabcolsep}{4pt}
\caption{Pretraining stages. Checkpoint steps are counted separately within each stage.}
\label{tab:training_schedule}
\begin{tabular}{llcr}
\toprule
Stage & Resolution & Text-token cap & Training step \\
\midrule
1 & 256-pixel level & 320 & 2,100,000 \\
2 & 512, then 256/512 & $512\!\to\!1024\!\to\!2048$ & 2,100,000 \\
3 & Mixed, area $\leq1024^2$ & 4096 & 370,000 \\
4 & Mixed, up to ${\sim}2{\times}1024^2$ & 4096 & 520,000 \\
\bottomrule
\end{tabular}
\end{table}

\paragraph{Caption sampling ratios.}
We mix original captions with multi-granularity recaptions. In the standard caption branch, original captions and recaptions are sampled at a 20:80 ratio when both are available, with uniform selection among valid recaption variants (1:1:1 for short/medium/long when all three are present). Missing alternatives fall back to available descriptions. We additionally annotate a selected subset of high-quality data with layout descriptions, serialized as bbox JSON, to provide explicit text and spatial supervision. These structured annotations are sampled according to domain-specific policies and annotation availability. Appendix~\ref{app:caption_sampling} provides the details.

\insight{Detailed annotations raise the performance ceiling; diverse annotations preserve generalization.}

\paragraph{RoPE design.}
Preliminary ablations (Appendix~\ref{app:rope_ablation}) show similar overall image-generation performance across RoPE coordinate assignments, with trade-offs across metrics. We retain a dedicated reference-index axis $r$ to allow future extensions to reference-conditioned generation. In the current setting without reference inputs, all tokens use $r=0$, so its associated feature subspace remains unrotated, 
while the other axes encode sequence and spatial positions. This supports more references without changing the positional encoding.


\paragraph{Muon and convergence efficiency.}
We choose hybrid Muon--AdamW for its sustained convergence advantage, applying Muon to eligible matrices and AdamW to the remaining parameters throughout pretraining. Our 1B ablation shows lower denoising loss in both early and late training, reaching losses of 0.41 and 0.403 with 33.2\% and 28.6\% fewer steps than AdamW, respectively. 
Appendix~\ref{app:optimizer_comparison} reports the convergence curves and measurement protocol; final training settings are given in Appendix~\ref{app:training_recipe}.

\section{Agentic Prompt Enhancement}
\label{sec:pe}

A concise user request leaves much of the literal content and visual organization unspecified. Our agentic prompt enhancer resolves these choices before rendering, turning user intent into content and layout conditions that the generator can execute. Our \emph{Layout PE} uses GPT-6 with modular task instructions at inference time.

\paragraph{Intent-specific task routing.}
GPT-6 selects task modules according to the requested medium. A shared core defines constraints, the JSON interface, and validation, while task-specific modules supply content and design guidance. Modules can be extended or customized without changing the interface or pretrained DiT.

\paragraph{Knowledge and visual context.}
The agent retrieves relevant knowledge and inspects visual references as needed to develop the requested content and composition. It adapts the specification to the target canvas. Reference images guide this planning process without directly conditioning the DiT.

\paragraph{Executable layout planning.}
The enhancer specifies literal text, elements, bounding boxes, typography, and spatial relationships in JSON. It checks user constraints, completeness, layout consistency, and text fit before passing the specification to the shared text-conditioning interface in Section~\ref{sec:conditioning}. This connects intent-level planning to the renderer's text and spatial control. Sections~\ref{sec:prompt_ablation} and~\ref{sec:backbone_comparison} compare prompts and backbones; Appendix~\ref{app:pe} covers implementation and skill rules.

\begin{table*}[t]
\centering
\begin{minipage}{\linewidth}
\centering
\begin{minipage}[t]{0.48\linewidth}
\centering
\small
\caption{Benchmark size and target-text length (characters).}
\label{tab:text_benchmark_comparison}
{\footnotesize
\setlength{\tabcolsep}{3pt}
\renewcommand{\arraystretch}{1.0}
\begin{tabular}{@{}lrr@{}}
\toprule
 & LongTextBench & \shortstack{\llb (ours)} \\
\midrule
EN / ZH cases & 160 / 160 & 120 / 120 \\
Total chars. & 38,489 & 321,545 \\
Mean / Median & 120.3 / 91 & 1,339.8 / 1,250 \\
Min. / Max. & 16 / 482 & 444 / 2,863 \\
\bottomrule
\end{tabular}
}
\end{minipage}

\hfill
\begin{minipage}[t]{0.48\linewidth}
\raggedright
\small
\setlength{\tabcolsep}{2pt}
\renewcommand{\arraystretch}{1.0}
\caption{Layout control on 7Bench.}
\label{tab:7bench}
\footnotesize
\begin{tabular*}{\linewidth}{@{\extracolsep{\fill}}lcc}
\toprule
Model & $\mathrm{mIoU}_{\mathrm{det}}$ $\uparrow$ & AUC $\uparrow$ \\
\midrule
GPT Image 2 & 0.30 & 0.16 \\
Nano Banana Pro & 0.33 & 0.16 \\
Qwen Image 3.0 Pro & 0.31 & 0.18 \\
Seedream 5.0 Pro & 0.33 & 0.18 \\
Ideogram 4.0 & 0.44 & 0.20 \\
\ours & \textbf{0.88} & \textbf{0.70} \\
\bottomrule
\end{tabular*}
\end{minipage}
\par\smallskip
\begin{minipage}[t]{\linewidth}
\centering
\scriptsize
\caption{\llb coverage (\%). Scores are weighted by target-unit count within each group; Mean-6 averages the six groups equally. See Appendix~\ref{app:longlongtextbench} for details.}
\label{tab:longlongtextbench}
\resizebox{\linewidth}{!}{
\begin{tabular}{lccccccc}
\toprule
Model &
EN-S $\uparrow$ &
EN-M $\uparrow$ &
EN-L $\uparrow$ &
ZH-S $\uparrow$ &
ZH-M $\uparrow$ &
ZH-L $\uparrow$ &
Mean-6 $\uparrow$
\\
\midrule

\multicolumn{8}{c}{\textbf{Closed-source Models}}\\
\midrule

Seedream 5.0 Pro
&92.8
&85.8
&41.7
&69.8
&28.3
&13.3
&55.3
\\

Qwen Image 3.0 Pro
&89.8
&94.3
&96.2
&98.9
&97.9
&88.5
&94.3
\\

Nano Banana Pro
&98.8
&99.6
&\textbf{97.4}
&98.1
&94.2
&91.4
&96.6
\\

GPT Image 2
&\textbf{99.8}
&\textbf{99.8}
&97.3
&\textbf{99.4}
&\textbf{99.2}
&\textbf{98.2}
&\textbf{99.0}
\\

\midrule

\multicolumn{8}{c}{\textbf{Open-source Models}}\\
\midrule


FLUX.2
&34.1
&19.6
&4.8
&6.1
&3.2
&2.9
&11.8
\\


Krea 2 Raw
&61.2
&36.9
&21.6
&39.1
&18.2
&13.7
&31.8
\\

Z-Image Base
&51.4
&42.3
&24.9
&70.4
&31.8
&24.3
&40.8
\\

SenseNova U1
&69.9
&45.7
&22.8
&72.4
&33.9
&18.7
&43.9
\\

ERNIE Image
&69.5
&63.7
&48.1
&79.6
&43.2
&30.2
&55.7
\\

Ideogram 4.0
&44.9
&\textbf{97.9}
&\textbf{97.5}
&48.8
&59.9
&31.8
&63.5
\\

Qwen Image 2512
&88.8
&86.7
&48.1
&89.9
&44.4
&31.3
&64.9
\\

\ours
&\textbf{98.0}
&97.1
&97.1
&\textbf{96.3}
&\textbf{96.7}
&\textbf{92.4}
&\textbf{96.3}
\\

\bottomrule
\end{tabular}
}
\end{minipage}
\end{minipage}
\end{table*}

\section{Experiments}
\label{sec:experiments}

\subsection{Experimental Setup and Benchmarks}
We separate evaluation of the pretrained generator from evaluation of complete generation systems. LongTextBench~\citep{geng2025xomni} and \llb measure bilingual text rendering, while 7Bench~\citep{izzo20257bench} measures layout control. These evaluations use the \ours backbone without our agentic prompt enhancer. \rla evaluates complete systems on concise design requests. Baseline scores on LongTextBench are collected from public reports, whereas all results on 7Bench, \llb, and \rla are obtained through our own evaluation. Benchmark construction, generation settings, and evaluation protocols are specified in Appendix~\ref{app:bench}.

\paragraph{LongTextBench saturation.} 
The collected LongTextBench results cluster between 96.05 and 98.85 overall (Appendix~\ref{app:longtextbench_results}). We therefore consider LongTextBench close to saturation for current strong models under its text-coverage metric. Its targets average only 120.3 characters (Table~\ref{tab:text_benchmark_comparison}); near-ceiling scores on these relatively short targets offer limited discrimination of sustained long-text rendering ability. We retain the detailed results in the appendix and emphasize longer text conditions in our main evaluation.

\paragraph{\textbf{LongLongTextBench}\unskip.}
We construct 240 literary-text rendering tasks, with 120 each in English and Chinese and three body-length bins. Table~\ref{tab:text_benchmark_comparison} compares the two benchmarks: \llb averages 1,339.8 characters per target, approximately $11.1\times$ LongTextBench, and contains 321,545 characters in total, approximately $8.4\times$ as many. Its shortest target (444 characters) already approaches LongTextBench's longest (482), shifting the evaluation toward dense, page-length text. We reuse the LongTextBench OCR and word/character multiset-recall metric. Coverage is computed within each language--length group and averaged equally across the six groups (Mean-6). Construction and scoring details are given in Appendix~\ref{app:longlongtextbench}.

\paragraph{RealLayoutArena.}
We construct 61 tasks from manually selected design references: 20 posters, 20 infographics, and 21 user interfaces. Raters compare two outputs for the same short request, with model names hidden, considering instruction following, rendering quality, and spatial organization. We fit Bradley--Terry ratings on an Elo scale with asymptotic 95\% confidence intervals; ties and \emph{both bad} count as half-wins. Appendix~\ref{app:arena_protocol} details the protocol.

\subsection{Base DiT Evaluation}
\label{sec:base_eval}
We first evaluate capabilities acquired by the pretrained backbone independently of our prompt enhancement pipeline. These results characterize the resulting model; they do not isolate the causal contribution of individual pretraining choices.

\paragraph{Text rendering.}
We emphasize \llb to assess rendering under substantially longer text conditions (Table~\ref{tab:longlongtextbench}); supplementary LongTextBench results are reported in Appendix~\ref{app:longtextbench_results}. \ours obtains 96.3 Mean-6 on \llb, the highest score among the open-weight models listed in the table. Its scores on the longest English and Chinese conditions are 97.1 and 92.4, respectively. Text recall is high overall but lower for the longest Chinese texts.

\paragraph{Layout control.}
Table~\ref{tab:7bench} reports mean IoU over detected target objects ($\mathrm{mIoU}_{\mathrm{det}}$) and Layout AUC. \ours achieves 0.88 and 0.70, respectively, compared with 0.44 and 0.20 for Ideogram~4, leading all evaluated models on both metrics. These results demonstrate precise object placement through the text-conditioning interface. Appendix~\ref{app:7bench_protocol} details the evaluation protocol, filtered outputs, detection rates, and results including undetected objects.

\subsection{End-to-End Evaluation on \rla}
\label{sec:arena}
\begin{figure*}[t]
\centering
\includegraphics[width=\linewidth]{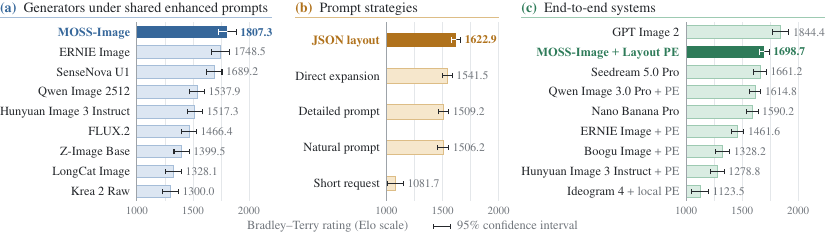}%
{\phantomsubcaption\label{fig:arena_dit}}%
{\phantomsubcaption\label{fig:arena_pe}}%
{\phantomsubcaption\label{fig:arena_main}}%
\caption{Human preference on \rla: (a) generators under shared enhanced natural-language prompts; (b) prompt strategies; and (c) end-to-end systems. Bars show Bradley--Terry ratings on an Elo scale with asymptotic 95\% confidence intervals. Ratings are fitted separately for the three comparison groups.}
\label{fig:arena_comparisons}
\end{figure*}

We compare nine end-to-end systems on the same short design requests.Figure~\ref{fig:arena_comparisons}\subref{fig:arena_main} reports the results. Appendix~\ref{app:arena_protocol} describes the evaluation protocol and configurations.

\paragraph{End-to-end results.}
\ours achieves 1698.70 Elo (95\% CI: 1655.96--1741.43), ranking second behind GPT Image~2 (1844.35) and ahead of Seedream~5.0 Pro (1661.16), Qwen Image~3.0 Pro with PE (1614.84), and Nano Banana Pro (1590.19). These results demonstrate competitive end-to-end performance on complex graphic-design requests.

\paragraph{Content quality and rendering fidelity.}
Arena judgments jointly assess instruction following, content completeness, text readability, and visual organization. This evaluation captures whether a system can turn a concise request into a coherent, usable design. Supplementary GPT-6 rubric-based evaluations are reported in Appendix~\ref{app:rubric_evaluation}.

\subsection{Ablation Studies}
\label{sec:ablations}
We examine prompt configurations for \ours and compare generators under shared enhanced natural-language prompts.

\subsubsection{Prompt Strategies}
\label{sec:prompt_ablation}
Figure~\ref{fig:arena_comparisons}\subref{fig:arena_pe} compares five strategies. \textit{Short} and \textit{Detailed} use original descriptions; \textit{Direct expansion} expands short requests without agentic routing or reference retrieval; \textit{Natural} and \textit{JSON layout} express Layout PE plans as prose and JSON. See Appendix~\ref{app:eval_configurations} for generation details.

Detailed prompts and direct expansion obtain 1509.21 and 1541.45 Elo, respectively, compared with 1081.70 for short requests. These results demonstrate the value of supplying explicit content and design instructions. Among the five configurations, JSON layout achieves the highest rating (1622.87), while the natural-language configuration scores 1506.20.

\subsubsection{Backbones under Shared Enhanced Natural Prompts}
\label{sec:backbone_comparison}
Figure~\ref{fig:arena_comparisons}\subref{fig:arena_dit} compares nine generators using the same natural-language rewrite of each Layout PE specification, without additional prompt enhancement. \ours achieves the highest rating at 1807.34 Elo (95\% CI: 1727.44--1887.24), followed by ERNIE Image (1748.51) and SenseNova-U1 (1689.23). These results show the backbone follows detailed text-and-layout instructions.

\section{Case Study}

\paragraph{Layout-Guided Zero-Shot Modification.}
\label{sec:layout_zero_shot}
Modifying the layout prompt enables targeted content replacement and element rearrangement while retaining the overall design. These changes are achieved through text-to-image regeneration without conditioning on the original image or additional training. Qualitative examples are provided in Appendix~\ref{app:layout_editing_cases}.

\paragraph{Line-Break Alignment in Text Annotations.}
\label{sec:linebreak_analysis}
We observe that prompts with explicit line breaks help the model establish the character content of each line earlier and more readily during denoising, leading to better final text rendering. Appendix~\ref{app:linebreak_cases} provides a qualitative comparison.

\paragraph{Bounding-Box Coordinate Representations.}
\label{sec:coordinate_analysis}
We observe that zero-centered absolute bounding-box coordinates generalize better to resolutions unseen during training than relative coordinates normalized to $[0,1000]$. Appendix~\ref{app:bbox_coordinate_cases} provides a qualitative comparison.

\section{Related Works}
\label{sec:related}
A detailed discussion of related work is provided in Appendix~\ref{app:related}.

\section{Limitations}
\label{sec:limitations}

The current \ours backbone supports text-to-image generation but does not directly condition on reference images or support image editing. Extending the model to image-conditioned generation and iterative design refinement would broaden its use in creative workflows; these capabilities remain future work under our current computational budget.


The prompt enhancer relies on external language models, whose limitations in aesthetic judgment, reference interpretation, and factual errors can reduce specification quality despite faithful rendering.

\section{Conclusion}
\label{sec:conclusion}

We present \ours, an open complex graphic generation system built around a 5B diffusion transformer pretrained from scratch. Its training recipe combines broad image-text data, multi-granularity and structure-aware supervision, and multi-stage resolution training. The resulting backbone supports strong bilingual long-form text rendering and layout control, while an agentic prompt enhancer translates concise requests into detailed generation specifications. \llb and \rla evaluate rendering capability and complete-system performance, complemented by prompt-strategy and shared-prompt cross-backbone comparisons.

\ours achieves state-of-the-art performance among open-weight models on \llb with a Mean-6 score of 96.3. On \rla, \ours ranks second in the end-to-end comparison, while its backbone achieves the highest rating under shared enhanced natural-language prompts. The prompt comparisons show that supplying detailed content and design instructions improves human preference over concise requests. Together, these results support combining a capable renderer with explicit generation specifications for complex graphics.

\section*{Contributors}

\textbf{Project Lead}

Zhe Xu$^{*}$, Donghua Yu$^{*}$, Qinyuan Cheng, Zhaoye Fei

\authornote{$^{*}$Equal Contribution.}

\textbf{Core Contributors}

Zhe Xu, Donghua Yu, Tengyue Zhang, Yanbing Zhang

\textbf{Contributors}

Ying Zhu, Yinze Luo, Qianyi Wu, Zhiyu Zhang, Wenbo Zhang, Yiyang Zhang, Ruixiao Li, Yiwei Zhao, Qian Chen, Haowei Zhang, Tianyi Liang, Chenghao Liu, Qi Luo, Zhijie Cao, Xiangyu Peng, Yang Gao, Shimin Li

\textbf{Advisor}

Xie Chen, Xipeng Qiu

\clearpage
\bibliographystyle{unsrtnat}
\bibliography{content/references}

\appendix
\appendixpage
\section{Architecture and Training Configuration}
\label{app:training_details}

\subsection{Model Configuration}
\label{app:model_configuration}
Table~\ref{tab:model_configuration} summarizes the diffusion generator and its frozen encoders. The 5B designation refers to the diffusion generator and excludes the frozen text encoder and VAE.

\begin{table}[htbp]
\centering
\small
\caption{Model configuration of \ours.}
\label{tab:model_configuration}
\begin{tabular}{@{}p{0.34\linewidth}p{0.62\linewidth}@{}}
\toprule
Component & Configuration \\
\midrule
Diffusion generator & 5B single-stream Transformer \\
Text encoder & Qwen3.5-9B (frozen) \\
Text features & Final-layer token hidden states, dimension 4096 \\
Text adapter & Linear projection $4096\rightarrow3072$; two text-only Transformer blocks \\
Joint backbone depth & 42 blocks, excluding the text adapter and encoder \\
Hidden size & 3072 \\
Attention & 24 heads, 128 dimensions per head \\
FFN hidden size & 8192 \\
Position axes & $(r,t,w,h)$: reference index, temporal position, width, height \\
RoPE configuration & Leading 32 of 128 head dimensions unrotated; remaining dimensions use $(t,w,h)$ \\
VAE & FLUX.2 VAE (frozen), loaded from FLUX.2-klein-4B \\
VAE latents & 32 channels; spatial downsampling factor 8 \\
Latent patch size & $2\times2$; 128 values per token before projection \\
Effective image stride & 16 pixels; token grid $(H/16)\times(W/16)$ \\
Final-stage text limit & 4096 tokens (Stage~3) \\
\bottomrule
\end{tabular}
\end{table}

\paragraph{Latent patchification.}
The FLUX.2 VAE downsamples each spatial dimension by a factor of 8. We group each $2\times2$ block of latent cells into one DiT input token, giving an effective image stride of 16 pixels. The number of image tokens is $N_{\mathrm{img}}=(H/16)(W/16)$: square images at $256^2$, $512^2$, and $1024^2$ yield 256, 1024, and 4096 image tokens, respectively.

\paragraph{Positional coordinates.}
The 4D coordinate system uses axes $(r,t,w,h)$ for reference index, sequence position, width, and height. Without reference inputs, all tokens use $r=0$, so the leading quarter of each attention head assigned to the reference axis remains unrotated, yielding partial RoPE. The remaining dimensions encode $(t,w,h)$. Spatial coordinates are centered at zero, and the image sequence position follows the text positions. The coordinate $t$ is a positional index rather than the diffusion timestep.

\paragraph{Sequence lengths and packing.}
The text-length limits increase across training stages as shown in Table~\ref{tab:training_schedule}. In Stage~3, a $1024^2$ image with a full-length text condition contains $4096+4096=8192$ image and text tokens. The multi-sample packing budget is 8576 tokens per pack, with a static execution length of 8588. These batching settings are distinct from the per-sample text limit.

\subsection{Training Configuration}
\label{app:training_recipe}
Pretraining proceeds through three stages with increasing resolution and text length. The final Stage~3 checkpoint is taken at step 310,000, using exponential moving average (EMA) weights. We describe the loss, noise distribution, optimization, and batching settings below.

\paragraph{Flow-matching loss.}
For image $i$, let $N_i$ be its number of image latent tokens and $C$ the token channel dimension. For packed training in the final stage, the loss is
\begin{equation}
\mathcal{L}_{\mathrm{FM}}=
\mathbb{E}\left[
\frac{1}{\sum_i N_i}\sum_i\sum_{p=1}^{N_i}
\frac{1}{C}\left\|
 v_\theta(z_{\sigma_i},1000\sigma_i,c_i)_p
 -(\epsilon_i-z_{0,i})_p
\right\|_2^2
\right].
\end{equation}
Text tokens do not contribute to this prediction loss, and there is no additional $\sigma$- or SNR-dependent weighting. Different-resolution images contribute in proportion to their valid image-token counts. Stage~2 transitions from per-sample to per-token normalization as the packed-training configuration is expanded.

\paragraph{Noise sampling.}
Table~\ref{tab:noise_schedule} records the four sampling intervals in Stage~3. The scheduler uses $T=1000$ time steps. For the final interval, define $Z\sim\mathcal{N}(\mu,s^2)$, $S=1-\operatorname{sigmoid}(Z)$, and
\[
F(x)=\Pr(S\leq x)
=1-\Phi\!\left(\frac{\log((1-x)/x)-\mu}{s}\right),
\]
where $\Phi$ is the standard normal CDF. With $\mu=-1.09$, $s=0.95$, cutoff $a=0.95$, and uniform-mixture weight $\rho=0.01$, the final discrete distribution is
\begin{equation}
\sigma_j=\frac{j}{T},\qquad
p_j=(1-\rho)\frac{F(aj/T)-F(a(j-1)/T)}{F(a)}
+\frac{\rho}{T},\quad j=1,\ldots,T.
\end{equation}
The cutoff is applied before rescaling onto the full grid $\{0.001,\ldots,1.000\}$. The final interval mixes truncated logit-normal sampling with 1\% uniform sampling. The first two intervals additionally sample the full-noise endpoint with probability 0.01, as shown in Table~\ref{tab:noise_schedule}.

\begin{table}[htbp]
\centering
\small
\setlength{\tabcolsep}{4pt}
\caption{Stage~3 noise-sampling schedule. The final interval mixes in 1\% discrete uniform sampling. The full-noise column gives the probability of separately selecting the full-noise endpoint.}
\label{tab:noise_schedule}
\begin{tabular}{rcccc}
\toprule
Stage~3 steps & Shift & $(\mu,s)$ & Cutoff & Full-noise override \\
\midrule
1--140,000 & 3 & $(0,1)$ & None & 0.01 \\
140,001--207,000 & 1 & $(-1.45,1.05)$ & None & 0.01 \\
207,001--274,000 & 1 & $(-1.45,1.05)$ & 0.95 & 0 \\
274,001--310,000 & 1 & $(-1.09,0.95)$ & 0.95 & 0 \\
\bottomrule
\end{tabular}
\end{table}

For untruncated sampling, $u=\operatorname{sigmoid}(Z)$ selects index $k=\min(\lfloor Tu\rfloor,T-1)$ in a descending noise grid. The initial interval uses a grid with shift 3; subsequent intervals use shift 1 and the logit-normal parameters listed in Table~\ref{tab:noise_schedule}.

\paragraph{Optimizer and numerical precision.}
All three pretraining stages use a hybrid Muon--AdamW optimizer. In the final configuration, eligible two-dimensional matrices are updated with Muon; non-matrix, embedding-like, output-head, and explicitly excluded parameters use AdamW. Fused QKV and gated-FFN matrices are split for Muon updates. Table~\ref{tab:optimizer_configuration} gives the principal settings. Weight decay is disabled for bias, normalization, one-dimensional/scalar, and embedding parameter groups. A separate decay-to-one rule with coefficient 0.01 is applied to the q/k normalization weights in the final Transformer block.

\begin{table}[htbp]
\centering
\small
\caption{Optimizer hyperparameters settings.}
\label{tab:optimizer_configuration}
\begin{tabular}{ll}
\toprule
Setting & Value \\
\midrule
Learning Rate (AdamW) & $5\times10^{-5}$ \\
Learning Rate (Muon) & $5\times10^{-5}$, RMS-matched\\
AdamW & $\beta_1=0.9$, $\beta_2=0.95$, $\varepsilon=10^{-8}$ \\
Muon & $\mu=0.95$, 5 N--S steps \\
Weight Decay & 0.01\\
Gradient Clipping & 1.0 \\
EMA & 0.9999\\
Learning-rate Warmup & 2000 steps \\
\bottomrule
\end{tabular}
\end{table}


\paragraph{Text formatting and packing.}
Caption, structural, and metadata fields are combined before applying the Qwen chat template and tokenization. The final-stage limit of 4096 tokens applies to the complete formatted condition. We use right padding and truncate conditions to this limit; attention masks identify valid text tokens for variable-length packing. Each device processes up to 8576 image-and-text tokens per microbatch. With 512 devices and two gradient-accumulation steps, the nominal capacity is 8,781,824 tokens per update. Actual image and valid-token counts depend on packing, while the static execution length of 8588 includes padding.




\section{Pretraining Dataset Details}
\label{app:data_cons}
We construct the pretraining corpus from multiple image--text sources and apply SSCD-based duplicate detection and an in-house AI-image detector during corpus preparation. Source-specific preprocessing, including the quality filters used by ReLAION, precedes this stage. We mix sources in their original proportions. Training captions are predominantly in English, without dedicated Chinese-caption annotation.

\subsection{Data Sources and Preprocessing}
\label{app:source_audit}
Table~\ref{tab:source_dedup_counts} summarizes the input and selection counts for the listed sources. Inputs are records received by the data pipeline, rather than the sizes of the original website crawls. The table reports source-level records.

\begin{table}[htbp]
\centering
\footnotesize
\setlength{\tabcolsep}{6pt}
\caption{Per-source record counts. The last column gives preprocessing output sizes.}
\label{tab:source_dedup_counts}
\begin{tabular}{lrr}
\toprule
Source & Input records & Retained records \\
\midrule
ReLAION~\citep{relaion} & $1.95\times 10^{9}$ & $1.19\times 10^{9}$ \\
OmniCorpus~\citep{li2024omnicorpus} & $3.21\times 10^{9}$ & $1.82\times 10^{9}$ \\
datacomp~\citep{gadre2023datacomp} & $2.91\times 10^{8}$ & $1.67\times 10^{8}$ \\
medtrinity\_25m~\citep{xie2024medtrinity25mlargescalemultimodaldataset} & $1.85\times 10^{7}$ & $1.36\times 10^{7}$ \\
optical-rs~\citep{selectivemae} & $1.32\times 10^{7}$ & $7.35\times 10^{6}$ \\
marqo~\citep{zhu2024generalized} & $9.81\times 10^{6}$ & $4.21\times 10^{6}$ \\
git\_10m~\citep{Text2Earth} & $9.81\times 10^{6}$ & $9.72\times 10^{6}$ \\
megalith~\citep{boerbohan2024megalith} & $8.56\times 10^{6}$ & $8.15\times 10^{6}$ \\
commoncatalog ~\citep{gokaslan2023commoncanvas}& $7.89\times 10^{6}$ & $7.78\times 10^{6}$ \\
text\_atlas~\citep{wang2025textatlas5m} & $5.40\times 10^{6}$ & $2.87\times 10^{6}$ \\
humancaption~\citep{dai2024humanvlmfoundationhumanscenevisionlanguage} & $4.90\times 10^{6}$ & $1.14\times 10^{6}$ \\
textrecap~\citep{chestnutlzj_textrecap} & $2.42\times 10^{6}$ & $1.38\times 10^{6}$ \\
google\_landmarks ~\citep{noh2017largescale}& $1.26\times 10^{6}$ & $1.21\times 10^{6}$ \\
wikiart~\citep{saleh2015largescale} & $8.14\times 10^{4}$ & $7.67\times 10^{4}$ \\
place365~\citep{zhou2017places} & $3.65\times 10^{4}$ & $3.46\times 10^{4}$ \\
poster100k ~\citep{chen2025postercraft}& $7.09\times 10^{3}$ & $6.37\times 10^{3}$ \\
\bottomrule
\end{tabular}
\end{table}

The effect of additional heuristic filtering is studied separately in Appendix~\ref{app:filtering_study}.

\subsection{Preliminary Filtering Experiments}
\label{app:filtering_study}

\paragraph{Experimental setup.}
We conducted an early comparison between a deduplicated, recaptioned ReLAION baseline and an additionally filtered dataset produced by the \texttt{pretrain\_v1} pipeline. Here, ``unfiltered'' denotes omission of this additional filter, rather than raw data without preprocessing. Both runs use the same 24-layer multi-stream diffusion transformer (hidden width 1280, feed-forward width 3072, and 20 attention heads), Qwen3.5-0.8B text encoder, and ViViT2-REPAE autoencoder. Training uses $256\times256$ images, 32 GPUs, a global batch size of 1024, a learning rate of $10^{-4}$, and recaption probability 0.8. Each run reaches 1.6M steps, with 32 recorded checkpoints at 50K-step intervals. These experiments use an early model configuration rather than the final 5B backbone.

\paragraph{Filtering criteria and retained data.}
The additional filter applies the criteria in Table~\ref{tab:data_filtering_criteria} conjunctively. A NULL value passes only its corresponding field-level check; every other applicable condition must still be satisfied. Of 1,183,622,445 source records, the pipeline retains 513,110,382 (43.35\%).

\begin{table}[htbp]
\centering
\small
\caption{Additional data-filtering criteria. All conditions must hold for non-NULL fields; NULL values pass the corresponding check.}
\label{tab:data_filtering_criteria}
\begin{tabular}{ll}
\toprule
Field & Retention condition \\
\midrule
Width and height & Each $\geq 192$ pixels \\
Aspect ratio & $\geq 0.33$ and $<3$ \\
Entropy & $\geq 4$ \\
Brightness & $\geq 0.1$ and $<0.9$ \\
Saturation & $<0.8$ \\
Sum of watermark area ratios & $<0.1$ \\
Natural score & $\geq 0.2$ \\
\bottomrule
\end{tabular}
\end{table}

\paragraph{Evaluation protocol.}
We evaluate on DPG-Bench, GenEval, OneIG, and FDR6-LAION. FDR6-LAION measures distributional image quality using the six-representation normalized Fr\'echet metric FDr$^6$~\citep{yang2026representation}. For each feature space, Fr\'echet distance (FD) compares the means and covariances of generated-image features with real-image reference statistics. We aggregate six feature spaces: Inception-v3, CLIP, ConvNeXt-v2, DINOv2, MAE, and SigLIP2. The score is $\frac{1}{6}\sum_{i=1}^{6}\mathrm{FD}_i/c_i$, with fixed normalization constants $c_i=(1.68,5.60,56.87,14.19,0.04,0.60)$ in the same order. This normalization accounts for the different numerical scales of the feature spaces; lower scores indicate closer agreement with the reference distribution. Each reported checkpoint score is computed from 50,000 generated images. We sweep classifier-free guidance (CFG) scales $\{1,1.25,1.5,1.75,2,3,4,5\}$ and compute the six-space aggregate separately at each scale. For each run and checkpoint, we report the minimum aggregate score over this sweep. Both final checkpoints at 1.6M training steps attain their minimum at CFG~3.

\begin{figure}[htbp]
\centering
\includegraphics[width=\linewidth]{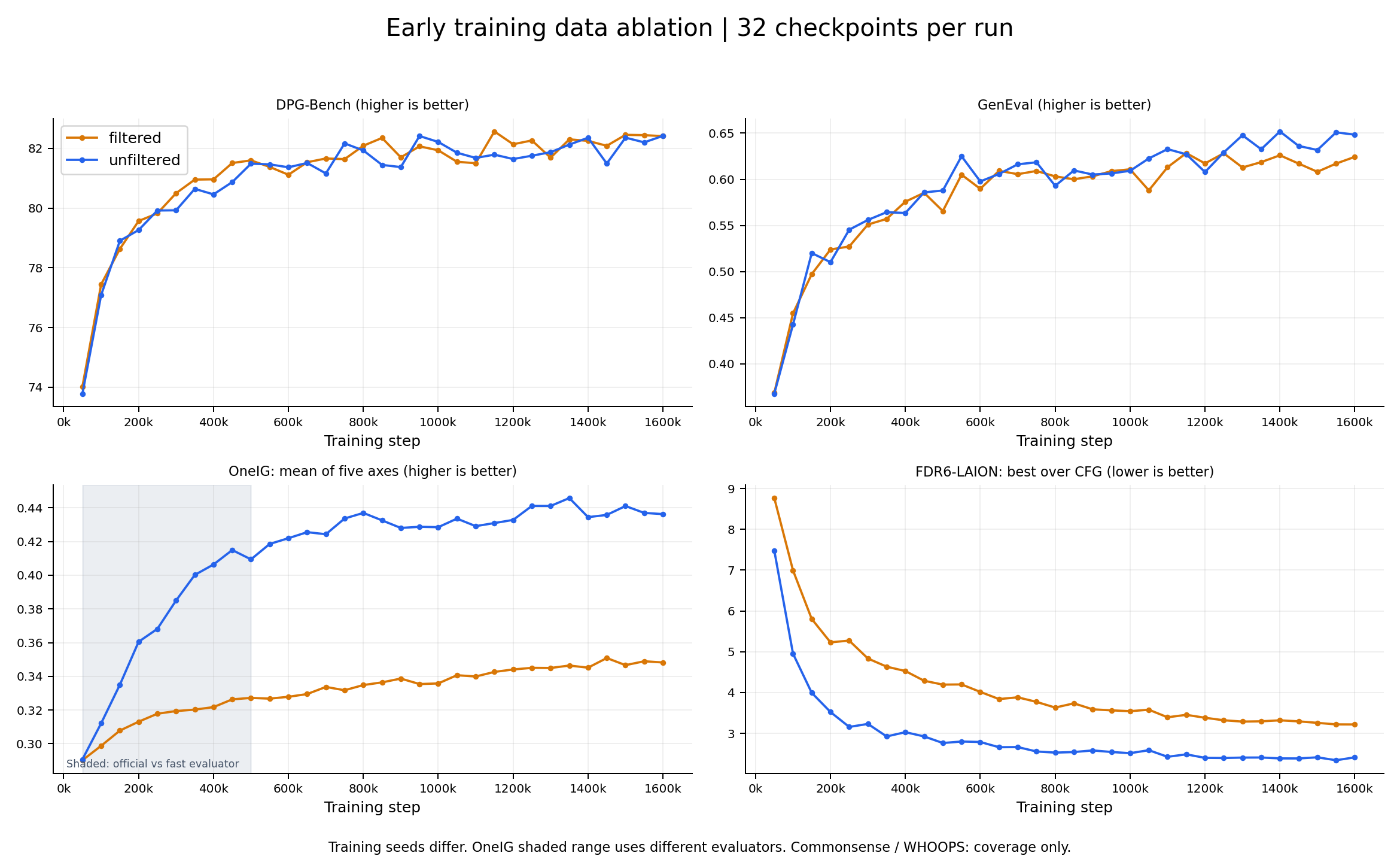}
\caption{\textbf{Early data-filtering comparison.} Curves show recorded checkpoint scores without smoothing. FDR6-LAION is the minimum over the recorded CFG sweep. Each curve represents one training run.}
\label{fig:data_filtering_comparison}
\end{figure}


\begin{table}[htbp]
\centering
\small
\caption{Results at 1.6M training steps. FDR6-LAION reports the minimum over the recorded CFG sweep.}
\label{tab:data_filtering_comparison}
\setlength{\tabcolsep}{5pt}
\begin{tabular}{lrrrr}
\toprule
Training data & DPG-Bench $\uparrow$ & GenEval $\uparrow$ & OneIG $\uparrow$ & FDR6-LAION $\downarrow$ \\
\midrule
Baseline & 82.420 & 0.6484 & 0.4363 & 2.4126 \\
Additional filtering & 82.411 & 0.6244 & 0.3482 & 3.2184 \\
\bottomrule
\end{tabular}
\end{table}

\paragraph{Results and implications.}
Additional filtering degrades overall generation quality in this ablation. At 1.6M steps, the filtered model performs worse on GenEval, OneIG, and FDR6-LAION, while DPG-Bench remains nearly unchanged (Table~\ref{tab:data_filtering_comparison}). The filtered model also has a higher (worse) FDR6-LAION score at all 32 evaluated checkpoints (Figure~\ref{fig:data_filtering_comparison})l. However, FDr$^6$ may not provide a fair comparison when training datasets differ, so these results should be interpreted only as indicative of trends. Text rendering is particularly affected: the final OneIG text score drops from 0.5243 to 0.1180 after filtering. We therefore retain broad training-data coverage and omit this additional heuristic filter.

\section{Caption Construction and Examples}
\label{app:caption}

\subsection{Caption Generation and Sampling}
\label{app:caption_sampling}
We generate short, medium, and long descriptions to provide multi-granularity textual supervision, and sample among the available captions during training.

\paragraph{Generation recipe.}
We generate three complementary descriptions for each image: a short single-sentence summary, a medium paragraph condensed from the long description, and a detailed account of composition, subjects, actions, environment, and style. The three levels differ in semantic detail rather than fixed token or word counts. Caption generation is conditioned on the image, its original caption, and a descriptive perspective. Visible text is transcribed without translation or transliteration, preserving line breaks, reading order, location, alignment, font, color, and spacing.

\paragraph{Availability and observed lengths.}
We characterize caption lengths using 5,792 images sampled across 94 data shards, with up to 64 uniformly spaced images per shard. Of these, 4,335 contain a complete short/medium/long triplet. Table~\ref{tab:caption_lengths} reports token-length statistics for these triplets using the Qwen3.5-9B tokenizer, excluding special tokens, training prefixes, and layout annotations, and without truncation. The longest sampled long caption contains 3,126 tokens. The statistics describe this annotation sample.

\begin{table}[htbp]
\centering
\small
\caption{Token lengths of 4,335 sampled short/medium/long caption triplets, measured with the Qwen3.5-9B tokenizer.}
\label{tab:caption_lengths}
\begin{tabular}{lrrrr}
\toprule
Variant & Mean & P10 & Median & P90 \\
\midrule
Short & 20.86 & 16 & 20 & 26 \\
Medium & 76.61 & 59 & 75 & 97 \\
Long & 235.12 & 153 & 193 & 293 \\
\bottomrule
\end{tabular}
\end{table}

\paragraph{Conditional sampling.}
When both an original caption and a generated caption are available, we select the generated caption with probability 0.8; otherwise, we use the available source. We then sample uniformly among its nonempty descriptions, assigning probability $1/3$ to each length when all three are present. 

\subsection{Annotation Validation}
\label{app:caption_validation}
Caption validation checks JSON syntax and the presence of three nonempty descriptions. Layout validation checks required fields, four-integer bounding boxes, coordinate ranges, positive box area, unique element identifiers, parent references, and color formats. The Gemma- and Qwen-based annotation generators allow up to three retries for invalid outputs. These checks enforce the annotation structure; the coordinate transformations and field selection used for textual conditioning are described in Appendix~\ref{app:conditioning}.

\subsection{Conditioning Schemas and Coordinate Conventions}
\label{app:conditioning}
The structured annotation pipeline supports two main schema families: scene/element and summary/items. Both combine global descriptions with localized content and bounding boxes. Figure~\ref{fig:layout_schemas} illustrates the two formats; a supplementary format appends item-level annotations to a natural-language caption.

\paragraph{Scene/element schema.}
The scene/element format combines a scene category and global brief with composition metadata, localized elements, spatial anchors, and a palette. Object elements describe identity and appearance; text elements specify visible text, typography, and watermark status. The stored boxes use $[y_1,x_1,y_2,x_2]$ coordinates normalized to $[0,1000]$. Training transforms boxes to match the image geometry and samples normalized, absolute-pixel, or center-relative pixel coordinates with probabilities $0.4/0.3/0.3$ within this branch. The normalized cases are split equally between implicit and explicit coordinate-format markers. For a transformed image of height $H$ and width $W$, center-relative pixel boxes are
\[
[y_1-H/2,\;x_1-W/2,\;y_2-H/2,\;x_2-W/2],
\]
where $(x,y)$ are pixel coordinates after geometric transformation. Optional fields are dropped and element order is randomized, while \texttt{gen\_brief} and \texttt{elements} are retained (Table~\ref{tab:layout_policy}).

\paragraph{Summary/items schema.}
The summary/items format contains a global \texttt{summary} and localized \texttt{items}, labeled as panels, objects, characters, text, or watermarks. We convert provider-specific box conventions to normalized $[0,1000]$ $yxyx$ coordinates and transform them consistently with image resizing, cropping, and flipping. Fully cropped or degenerate items are removed. Element identifiers and parent references are discarded, leaving the localized descriptions and spatial coordinates as the structural condition. Summary fields and item attributes undergo the stochastic transformations in Table~\ref{tab:layout_policy}. The serialized condition is processed by the shared text encoder.

\begin{figure}[htbp]
\centering
\begin{minipage}[t]{0.48\linewidth}
\footnotesize
\textbf{(a) Scene/element}
\begin{verbatim}
{
  "scene_type": "poster",
  "gen_brief":
    "A red circle below SALE.",
  "layout": {
    "formula": "centered",
    "camera": "front"
  },
  "elements": [
    {"type": "obj",
     "salience": "primary",
     "identity": "circle",
     "shape": "solid disk",
     "colors": [{"color": "red",
            "hex": "#FF0000"}],
     "bbox": [300, 250, 800, 750]},
    {"type": "text",
     "text": "SALE",
     "where": "upper",
     "bbox": [80, 250, 200, 750],
     "shape": "bold black type",
     "is_watermark": false}
  ],
  "anchors":
    ["Title above circle."],
  "palette": ["#FFFFFF", "#FF0000"]
}
\end{verbatim}
\end{minipage}\hfill
\begin{minipage}[t]{0.48\linewidth}
\footnotesize
\textbf{(b) Summary/items}
\begin{verbatim}
{
  "summary": {
    "meta": ["poster"],
    "canvas": "single",
    "caption":
      "A red circle below SALE."
  },
  "items": [
    {"bbox": [300, 250, 800, 750],
     "label": "object",
     "caption":
       "a solid red circle",
     "tags": ["circle"],
     "palette": ["#FF0000"]},
    {"bbox": [80, 250, 200, 750],
     "label": "text",
     "text": "SALE",
     "caption": "bold black type",
     "tags": ["title"]}
  ]
}
\end{verbatim}
\end{minipage}
\caption{Illustrative scene/element and summary/items conditions in normalized $yxyx$ coordinates. The examples use schematic content and coordinates. Element identifiers and parent references are removed during preprocessing; optional attributes are subject to field dropout.}
\label{fig:layout_schemas}
\end{figure}

\begin{table}[htbp]
\centering
\small
\setlength{\tabcolsep}{4pt}
\caption{Stochastic preprocessing of structured annotations. Probabilities are conditional on the corresponding schema branch and the availability of the relevant fields.}
\label{tab:layout_policy}
\begin{tabular}{@{}p{0.17\linewidth}p{0.25\linewidth}p{0.52\linewidth}@{}}
\toprule
Schema & Operation & Policy \\
\midrule
Scene/element & Field dropout & Activate with $p=0.5$; conditional on activation, apply dropout to each configured field group with $p=0.5$. Retain \texttt{gen\_brief} and \texttt{elements}. \\
Scene/element & Ordering and aliases & Randomize element order. With $p=0.5$, use an alternative key for the appearance description. \\
\midrule
Summary/items & Structural fields & Remove \texttt{id} and \texttt{parent} with $p=1$. Drop the entire \texttt{summary} with $p=0.1$. \\
Summary/items & Summary attributes & Within a retained summary, drop \texttt{meta} with $p=0.1$, and \texttt{canvas}/\texttt{caption} with $p=0.5$ each. Replace a retained caption with the selected base prompt with $p=0.5$. \\
Summary/items & Item attributes & Apply $p=0.5$ dropout to palettes and other optional fields. Use three equal-weight modes for \texttt{tags}/\texttt{caption}: both, tags only, or caption only. \\
Summary/items & Element order & Shuffle items with $p=0.5$. \\
\bottomrule
\end{tabular}
\end{table}

\paragraph{Supplementary item-level conditions and sampling.}
A supplementary condition appends item-level boxes, labels, tags, and captions to a base natural-language description, together with a coordinate-system marker. This branch samples normalized, absolute-pixel, and center-relative pixel coordinates with probabilities $0.2/0.4/0.4$.  The inference-time specifications produced by the prompt enhancer are described in Appendix~\ref{app:pe}.

\clearpage
\section{RoPE Ablation and Reference-Axis Design}
\label{app:rope_ablation}

\paragraph{Coordinate assignments.}
We compare four positional-coordinate designs in preliminary 1B experiments: the baseline RoPE of the experimental backbone; modality separation along the sequence axis; modality separation along the $r$ axis; and sequential text--image coordinates with $r=0$. In the sequence-axis variant, text and image tokens occupy separate sequence-coordinate ranges while sharing $r=0$. In the $r$-axis variant, text and image tokens use $r=0$ and $r=1$, respectively. The sequential variant keeps $r=0$ for all tokens and places image sequence positions after the text positions, preserving the $r$ axis for reference indexing.

\paragraph{Evaluation.}
We evaluate DPG-Bench, GenEval, and OneIG; OneIG is the arithmetic mean of its alignment, text, reasoning, style, and diversity scores. Table~\ref{tab:rope_ablation} compares all four designs at 200K training steps and summarizes the longer comparison of the baseline and two modality-separation variants using ten checkpoints from 800K to 1.25M steps at 50K-step intervals. Each comparison uses the same generation scheduler.

\begin{table}[htbp]
\centering
\small
\setlength{\tabcolsep}{7pt}
\caption{RoPE coordinate ablations. The upper block compares four designs at 200K steps; the lower block reports checkpoint means from 800K to 1.25M steps. Higher is better for all metrics.}
\label{tab:rope_ablation}
\begin{tabular}{lrrr}
\toprule
Coordinate assignment & DPG-Bench & GenEval & OneIG \\
\midrule
\multicolumn{4}{l}{\textit{Four-way comparison at 200K steps}} \\
Baseline & 81.94 & 0.6313 & 0.3483 \\
Modality on sequence axis & 81.41 & 0.6101 & 0.3588 \\
Modality on $r$ axis & 80.93 & 0.6280 & 0.3597 \\
Sequential text--image, $r=0$ & 81.89 & 0.6026 & 0.3605 \\
\midrule
\multicolumn{4}{l}{\textit{Mean over 800K--1.25M steps}} \\
Baseline & 83.54 & 0.6941 & 0.4157 \\
Modality on sequence axis & 83.56 & 0.6867 & 0.4170 \\
Modality on $r$ axis & 82.77 & 0.6782 & 0.4255 \\
\bottomrule
\end{tabular}
\end{table}

\paragraph{Design choice.}
The coordinate assignments yield similar overall image-generation performance with trade-offs across metrics. Sequential text--image coordinates retain nearly the same DPG-Bench score as the baseline at 200K steps, with higher OneIG and lower GenEval. The longer comparison likewise shows that using the $r$ axis for modality separation improves OneIG while reducing DPG-Bench and GenEval. We retain a dedicated reference-index axis for extensibility. Without reference inputs, setting $r=0$ makes the associated rotations identity transformations, leaving that feature subspace unrotated and yielding partial RoPE. Distinct reference indices can be introduced within the same coordinate structure when extending the model to reference-conditioned generation. Appendix~\ref{app:model_configuration} specifies the configuration used by \ours.
\clearpage
\section{Optimizer Ablation: Training-Loss Convergence}
\label{app:optimizer_comparison}
\paragraph{Setup.}
We compare AdamW with hybrid Muon--AdamW in early ReLAION training of a 1.016B generator at $256\times256$, using a frozen Qwen3.5-0.8B-Base encoder. The primary pair uses logit-normal timestep sampling (mean 0, standard deviation 1) and 64 text tokens throughout. Both runs use BF16, 32 samples/device, gradient accumulation 1, learning rate $10^{-4}$, 2000 warmup steps, cosine decay, and weight decay 0.01. AdamW uses $\beta=(0.9,0.95)$; Muon uses momentum 0.95, Nesterov updates, five Newton--Schulz steps, and RMS matching, while a subset of parameters retains AdamW updates.

\begin{figure}[htbp]
\centering
\includegraphics[width=\linewidth]{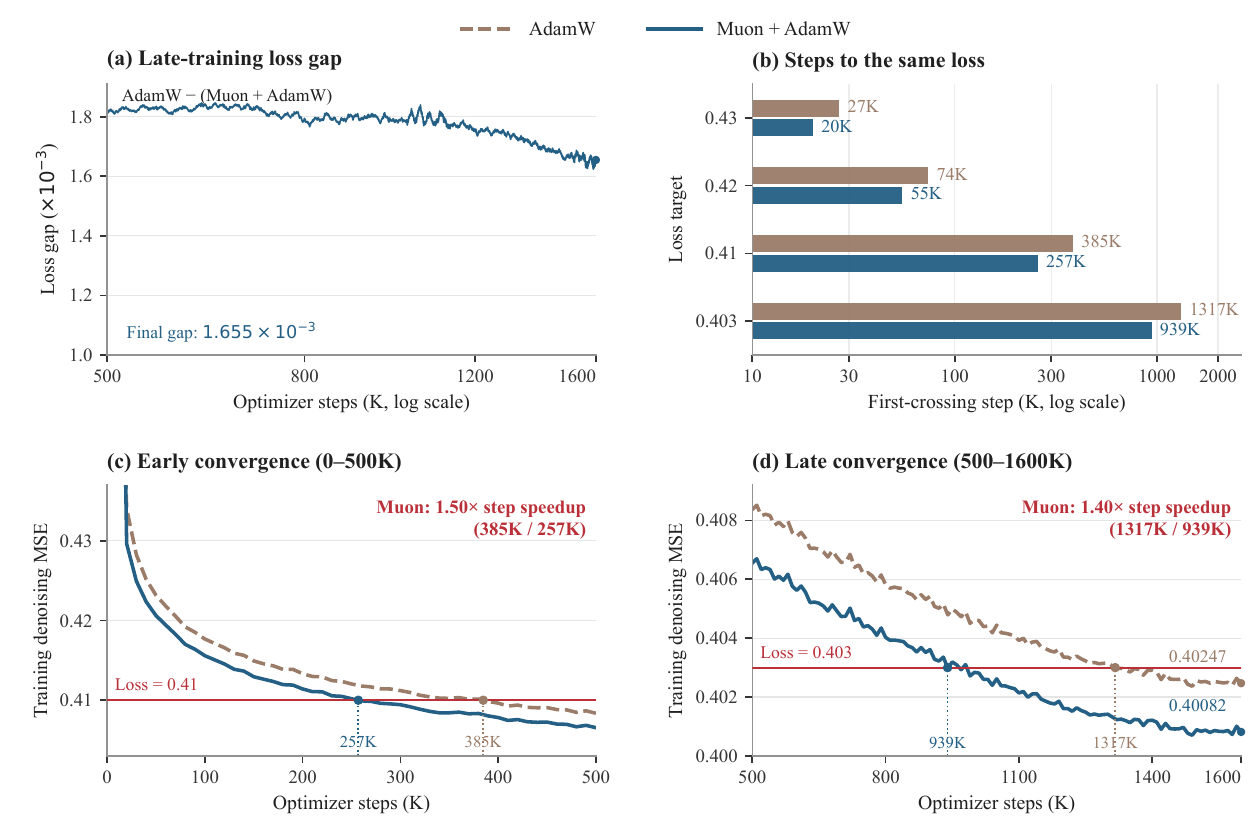}
\caption{\textbf{Muon--AdamW maintains a convergence advantage from early to late training.} Matched logit-normal runs: (a) the AdamW minus Muon--AdamW loss gap; (b) first-crossing steps at four loss targets; (c) early convergence through 500K steps; and (d) late convergence from 500K to 1.6M steps. Red lines and dots in (c,d) mark target losses and interpolated first crossings. Panels (a,b) use logarithmic step axes. Each curve represents one training run.}
\label{fig:optimizer_comparison}
\end{figure}

\paragraph{Loss measurement.}
We measure convergence using training denoising mean-squared error. Loss values recorded every 10 steps are averaged in non-overlapping 10K-step bins with at least 90\% coverage for Figure~\ref{fig:optimizer_comparison}(b--d). A target-loss step is the first intersection of the resulting piecewise-linear curve with that target, obtained by interpolation between adjacent bin endpoints. Step counts are rounded to the nearest 1K for display; ratios use the unrounded values. Panel (a) reports AdamW loss minus Muon--AdamW loss over 500K--1.6M steps, using 10K-step moving means evaluated every 100 steps.

\paragraph{Training-step efficiency.}
Muon--AdamW reaches loss 0.41 at 257K steps versus 385K for AdamW, requiring \textbf{33.2\% fewer updates} with an AdamW/Muon step ratio of $1.50\times$. This advantage persists later: at loss 0.403, the corresponding steps are 939K and 1317K, giving \textbf{28.6\% fewer updates} and a $1.40\times$ step ratio. The loss gap remains positive throughout 500K--1.6M steps, ending at $1.655\times10^{-3}$; the final 10K-step mean losses are 0.40082 for Muon--AdamW and 0.40247 for AdamW. This sustained convergence advantage motivates our use of hybrid Muon--AdamW throughout pretraining.

\paragraph{Sensitivity.}
Using 5K, 10K, or 20K bins, the first-crossing step ratios (AdamW/Muon) are 1.37, 1.50, and 1.47 at loss 0.41, and 1.35, 1.40, and 1.35 at loss 0.403, respectively. The convergence advantage therefore persists across aggregation widths in both early and late training.

\section{Prompt Enhancer Implementation}
\label{app:pe}

\subsection{GPT-6 Planner and Task Guidance}
We use GPT-6 as a planner with shared instructions for constraint extraction, text preservation, composition, readability, and structured output. Task-specific guidance covers posters, infographics, and user interfaces (Table~\ref{tab:pe_routes}). Routing prioritizes the medium explicitly requested by the user, followed by the inferred communication goal. For example, an educational poster follows the poster route unless an infographic is explicitly requested. User-specified content, literal strings, language, and canvas dimensions constrain the resulting plan.

\paragraph{Planner sessions.} The final Layout PE plans each task in an independent session. An earlier version planned all 61 \rla tasks within a single shared context. The main system uses the final version, while the prompt-strategy ablation and the shared-prompt generator comparison use the earlier version.

\begin{table}[htbp]
\centering
\small
\setlength{\tabcolsep}{4pt}
\caption{Task-specific planning guidance for the prompt enhancer. Suggested reference counts are adapted to the request and the available material.}
\label{tab:pe_routes}
\begin{tabular}{@{}p{0.16\linewidth}p{0.40\linewidth}p{0.38\linewidth}@{}}
\toprule
Route & Planning priorities & Reference guidance \\
\midrule
Poster & Information hierarchy, complete event or campaign content, typography, and coherent visual composition. & Search same-theme posters; typically inspect 3--5 distinct examples and compare content, density, and typography/layout. \\
Infographic & Develop factual explanations jointly with diagrams, labels, and reading order; preserve substantive content when grouping elements. & Inspect relevant visual references and consult authoritative sources for unfamiliar or uncertain facts. \\
User interface & Match the intended page, platform, viewport, version, and populated interface state. & Collect roughly 3--6 candidates and normally inspect at least 3 distinct screenshots; prefer 2 sources and 2 relevant page views when available. \\
\bottomrule
\end{tabular}
\end{table}

The planner is instructed to retrieve relevant material through web and image search and visually inspect candidate references. The guidance recommends 1--3 focused queries and 2--4 useful images, with task-specific guidance as listed in Table~\ref{tab:pe_routes}. For user interfaces, the search can be refined to match the requested page, platform, and populated state. Retrieved designs inform composition and appearance, while unfamiliar factual content is checked against authoritative sources.

For broad knowledge graphics, the infographic guidance suggests 6--9 substantive modules, 22--36 knowledge units, and approximately 300--500 visible English words on a $1024\times(1400\text{--}1600)$ canvas. The planner adjusts these targets to the requested scope, language, canvas, and readability. Related content is grouped into coherent visual units, with the number of elements determined by the design.

\subsection{Structured Prompt Format}
The planner returns one JSON object with eight root fields: \texttt{scene\_type}, \texttt{gen\_brief}, \texttt{aspect\_ratio}, \texttt{layout}, \texttt{elements}, \texttt{anchors}, \texttt{must\_not}, and \texttt{palette}. Compared with the representative scene/element training schema in Appendix~\ref{app:conditioning}, the inference format additionally includes \texttt{aspect\_ratio} and \texttt{must\_not}.

The global brief uses 60--120 English words; \texttt{layout} specifies composition and viewpoint. Elements use \texttt{type=obj} or \texttt{type=text}; object \texttt{salience} distinguishes primary and secondary roles. Type-specific fields encode appearance, literal copy, color, and placement. Every visible string is written literally in a text element, preserving the user's original characters and intended line breaks. Descriptive fields are in English. Related strings and mixed typography may share a box when their internal organization remains explicit. Up to three anchors encode important spatial relationships, and up to three \texttt{must\_not} entries describe concrete, request-specific visual exclusions.

The planner emits integer boxes in normalized $[0,1000]$ coordinates with a top-left origin and order $[y_{\min},x_{\min},y_{\max},x_{\max}]$. It requires positive box width and height and allows overlap when implied by the intended composition. For output dimensions $W\times H$, the planner checks text fit using the physical box dimensions $\Delta x\,W/1000$ and $\Delta y\,H/1000$, literal line breaks, and typography hierarchy. Boxes describe coherent visual units, not mandatory visible panels.

\subsection{Specification Validation}
Before returning a specification, the planner is instructed to check JSON syntax, required fields, coordinate validity, literal text, and consistency between the global composition and localized elements. Task-specific checks cover content completeness and readability. Intended line breaks are represented as escaped newline characters in JSON. Reference information is incorporated into self-contained visual descriptions, while research notes and source URLs are kept outside the image-description fields.

\section{Benchmark Construction and Evaluation}
\label{app:bench}

\subsection{\llb}
\label{app:longlongtextbench}

\paragraph{Tasks and source texts.}
The benchmark contains 240 tasks: 120 English and 120 Chinese, with 40 tasks per language in each of three text-length bins (Table~\ref{tab:longlong_construction}). We select 80 source entries, with 20 in each language--style group (English/Chinese and classical/modern). Each entry supplies three prefix excerpts, allowing length conditions to share source material. The English entries come from 40 Project Gutenberg works. Chinese classical entries include 17 excerpts from \emph{Mencius}, two from \emph{The Doctrine of the Mean}, and one from \emph{The Great Learning}, collected through the Chinese-poetry collection. Chinese modern entries comprise 20 excerpts from seven Lu Xun works on Chinese Wikisource, including 12 from \emph{The True Story of Ah Q}. Titles, authors, and source URLs accompany each entry.

\begin{table}[htbp]
\centering
\small
\setlength{\tabcolsep}{4pt}
\caption{\llb construction. Bin limits count body characters including retained whitespace and punctuation; observed ranges below exclude whitespace. Each bin has 40 tasks per language.}
\label{tab:longlong_construction}
\begin{tabular}{lrrr}
\toprule
Bin & Construction interval & Observed EN range & Observed ZH range \\
\midrule
Short & 180--420 & 335--355 & 402--420 \\
Medium & 700--1,200 & 962--1,013 & 1,164--1,200 \\
Long & 1,600--2,800 & 2,238--2,334 & 1,657--1,788 \\
\bottomrule
\end{tabular}
\end{table}

\paragraph{Construction and formatting.}
We clean source text, apply Unicode NFC normalization, remove control/format characters, and normalize whitespace. Chinese text is converted to simplified Chinese with OpenCC. Source passages are selected in a fixed order and truncated to prefixes; passages below the required minimum are rejected rather than repeated. Length bins are defined by character count, including whitespace and punctuation, with an additional lower bound on non-whitespace characters. Each input contains \texttt{A paper writes:}, title, author, and body. Automatic wrapping uses DejaVuSans font metrics for a $1024\times1024$ canvas with 88-pixel margins and searches body sizes from 32 down to 8 pixels, wrapping English by word and Chinese by character. These metrics construct the input line breaks without constraining the generator's actual font. The construction intervals count body text only; full-target statistics below also include the title and author.

\paragraph{Target-length statistics.}
Table~\ref{tab:text_benchmark_comparison} reports the scale of the full target strings, separately from the body-only construction intervals above. \llb contains 12 language--style--length slots with 20 tasks each, equivalently 40 tasks per language--length group. The complete targets contain 321,545 characters including whitespace, or 285,792 characters after removing whitespace. The full-target mean, median, and range are 1,339.8, 1,250, and 444--2,863 characters, respectively. These descriptive character counts are distinct from the normalized English-word and Chinese-character units used by the coverage metric.

Within the short bin, mean target lengths are approximately 468 English characters and 458 Chinese characters (roughly 460 overall). The medium-bin mean is approximately 1,244 characters pooled across languages. In the long bin, the means are approximately 2,848 for English and 1,770 for Chinese; the language-specific settings use a lower character limit for Chinese. By comparison, LongTextBench contains 320 tasks and 38,489 target characters, averaging approximately 184 per English task and 56.5 per Chinese task. These statistics distinguish rendering relatively short text within a scene from sustaining dense text across a page.

\paragraph{OCR and matching.}
We follow the X-Omni LongTextBench OCR and scoring procedure, using Qwen2.5-VL-7B-Instruct with a 4096-token output limit. Images are resized to $1024\times1024$ with LANCZOS interpolation. Ground truth drops the instruction prefix and retains title, author, and body. English preprocessing removes characters outside the scorer's whitelist, collapses whitespace, lowercases, and splits into words. Chinese preprocessing removes out-of-whitelist characters and whitespace and matches characters, without lowercasing embedded Latin letters. The whitelist covers basic Chinese characters, Latin letters, digits, and specified French accented characters; it excludes punctuation.

For image $i$, let $G_i$ and $O_i$ be the ground-truth and OCR multisets of matching units. We compute
\[
m_i=\sum_u\min\{\operatorname{count}_{G_i}(u),\operatorname{count}_{O_i}(u)\},
\qquad r_i=\frac{m_i}{|G_i|}.
\]
For language $\ell$ and length bin $b$, coverage and the overall score are
\[
C_{\ell,b}=\frac{\sum_{i\in(\ell,b)}m_i}{\sum_{i\in(\ell,b)}|G_i|},
\qquad \text{Mean-6}=\frac{1}{6}\sum_{\ell,b}C_{\ell,b}.
\]
Each generated image contributes once. Coverage is weighted by target-unit count within a group, followed by an equal-weight average over six groups. The metric does not check order or directly penalize extra OCR units; it is not edit distance or exact string accuracy. Pooling all matching units across languages yields a different micro-coverage score and mixes English words with Chinese characters.

\paragraph{Generation settings.}
For every evaluated model, we generate four images per task and include all four samples in scoring, without best-of-four selection. OCR inputs are resized to $1024\times1024$, while native generation resolutions follow each model's supported settings. Ideogram~4 safety-refusal images receive a score of zero.

\paragraph{Ideogram 4 refusals.}
\label{app:ideogram_refusals}
At CFG~4, Ideogram~4 produces 61 refusal images among 960 generation attempts, all in the EN-short group (61 of 160). These tasks feature relatively sparse text rendered in large type. No refusals occur in EN-medium, EN-long, or any of the 480 Chinese images. Table~\ref{tab:longlongtextbench} counts each refusal as zero coverage: its matched-unit count is set to $m_i=0$, while its target units remain in the denominator. We also evaluate a second convention that excludes refusal images from both the numerator and denominator. This leaves 99 scored images in EN-short and 160 in each of the other five groups, for 899 images in total.

Table~\ref{tab:ideogram_refusal_scoring} compares the two conventions. Mean-3 averages the three length-bin scores within a language, while language coverage and total coverage pool matching and target units over the corresponding images. Mean-6 retains equal weighting across the six language--length groups under both conventions. Excluding refusals raises EN-short coverage from 44.93\% to 72.41\% and Mean-6 from 63.47\% to 68.05\%; Chinese scores are unchanged.

\begin{table}[htbp]
\centering
\scriptsize
\caption{\llb coverage (\%) of Ideogram~4 at CFG~4 under two refusal-handling conventions.}
\label{tab:ideogram_refusal_scoring}
\resizebox{\linewidth}{!}{
\begin{tabular}{lccccccc}
\toprule
Refusal handling & EN-S $\uparrow$ & EN-M $\uparrow$ & EN-L $\uparrow$ & ZH-S $\uparrow$ & ZH-M $\uparrow$ & ZH-L $\uparrow$ & Mean-6 $\uparrow$ \\
\midrule
Scored as zero & 44.9 & 97.9 & 97.5 & 48.8 & 59.9 & 31.8 & 63.5 \\
Excluded & 72.4 & 97.9 & 97.5 & 48.8 & 59.9 & 31.8 & 68.1 \\
\bottomrule
\end{tabular}
}
\end{table}

With refusals scored as zero, EN Mean-3 and pooled coverage are 80.11\% and 92.34\%, respectively; excluding refusals raises them to 89.27\% and 96.00\%. ZH Mean-3 and pooled coverage remain 46.83\% and 44.03\% under both conventions. Total pooled coverage is 55.03\% with refusals scored as zero and 55.51\% with refusals excluded.

\subsection{\rla}
\label{app:arena_protocol}

\paragraph{Task construction.}
We construct 61 tasks from manually screened design references: 20 posters, 20 infographics, and 21 user interfaces. Each task is expressed as a concise Chinese request; the reference images contain text in multiple languages. Short, detailed, and enhanced prompts are alternative descriptions of the same task, allowing prompt strategies to be compared on a shared task set.

\paragraph{Generation and fixed outputs.}
A participant is a generator paired with a prompt strategy. Each participant generates a single image per task, using a fixed seed where the generator supports it, without re-sampling, manual filtering, or best-of-$k$ selection, and this image is held fixed for all raters. Target canvases preserve the reference aspect ratio at approximately one megapixel, with dimensions divisible by 16. The 31 target sizes span 1,034,752--1,063,424 pixels. Generation sizes follow model-specific constraints: Krea~2 Raw limits the longest side to 1024 pixels; HiDream-O1 and SenseNova U1 may produce larger images; Nano Banana Pro and Hunyuan Image~3 use their supported size and aspect-ratio options.

\paragraph{Presentation and judging.}
Raters compare two images for the same task while viewing its original short request. The interface hides model names and expanded prompts. Image pairs are fixed before voting, and left/right positions are randomized for each rater and pair. Pair selection uses participant- and comparison-group exposure weights, with a shuffled presentation order for each rater. The number of judgments can vary across pairs.

The choices are A better, B better, tie, both bad, and skip. Instructions prioritize adherence to the short request, then rendering quality, and then spatial relationships when content and quality are comparable. Raters consider completeness, readable and intact text/elements, grouping, reading order, and visual hierarchy. Optional defect tags provide supporting annotations rather than separately scored dimensions. Skip is reserved for images that cannot be loaded or judged.

\paragraph{Votes and Bradley--Terry ratings.}
 Wins and losses count as one and zero; tie and both-bad votes split one win equally. We fit
\[
P(i\succ j)=\frac{\exp(\theta_i)}{\exp(\theta_i)+\exp(\theta_j)},
\qquad E_i=1500+\frac{400}{\ln 10}\theta_i.
\]
Each participant receives 0.25 pseudo-wins and 0.25 pseudo-losses against a strength-one virtual anchor to keep estimates finite. Ratings are ordered by fitted $E_i$; they are not sequential Elo updates. Preference score rate, when reported, is $(w+0.5t)/(w+l+t)$, with both-bad included in $t$. Unequal opponent exposure makes this a different statistic from the fitted rating. Fitting aggregates votes without equal-task/category averaging or inverse-exposure weighting.

\paragraph{Confidence intervals.}
Asymptotic 95\% confidence intervals use the Fisher information of real comparisons, excluding pseudo-match curvature. Within each connected comparison component, we impose a zero-sum parameter constraint and take a generalized inverse to obtain standard errors, reporting $E_i\pm1.959964\,\mathrm{SE}(E_i)$. These intervals quantify uncertainty in relative ratings under the comparison model and do not adjust for correlations between votes from the same task or rater. We fit each comparison group separately using votes between its participants.

\paragraph{Evaluation coverage.}
\label{app:arena_snapshot}
Table~\ref{tab:arena_snapshot_coverage} summarizes the complete comparison groups used to fit the ratings. Figure~\ref{fig:arena_comparisons} presents nine end-to-end systems, five prompt strategies, and nine shared-prompt generators from these groups. Ratings use all valid within-group comparisons. Every model pair in each group receives at least one judgment.

\begin{table}[htbp]
\centering
\small
\setlength{\tabcolsep}{4pt}
\caption{Human-evaluation coverage of the rating groups on the 61 \rla tasks. A task--pair is one task paired with an unordered pair of participants. Coverage counts task--pairs with at least one valid vote.}
\label{tab:arena_snapshot_coverage}
\begin{tabular}{lrrrrr}
\toprule
Group & Participants & Votes & Model pairs & Task--pairs & Coverage \\
\midrule
Main systems & 11 & 1420 & 55/55 & 1045/3355 & 31.1\% \\
Prompt strategies & 9 & 1533 & 36/36 & 848/2196 & 38.6\% \\
Shared natural prompts & 10 & 576 & 45/45 & 539/2745 & 19.6\% \\
\bottomrule
\end{tabular}
\end{table}

Skip and rejected votes are excluded, and all remaining votes receive equal weight. In the main-system comparison, \ours with Layout PE receives 293 wins, 100 ties, and 64 losses across 457 votes, giving a preference score rate of 75.1\%. In the shared-prompt comparison, \ours receives 98 wins, 18 ties, and 12 losses across 128 votes, giving a preference score rate of 83.6\%. The Bradley--Terry ratings additionally account for opponent strength.

\subsection{Baseline and Ablation Configurations}
\label{app:eval_configurations}
\paragraph{End-to-end systems.}
The main \ours system uses the final Layout PE, with one independent planner session per task (Appendix~\ref{app:pe}). The eight baseline systems receive the original short request and use their respective prompt-processing pipelines. The Ideogram~4 baseline uses local Qwen3VL-8B prompt enhancement. The comparison evaluates the complete generation pipeline for each configuration.

\paragraph{Prompt configurations.}
\textit{Short} uses the concise task request, \textit{Detailed} uses the benchmark's detailed description, and \textit{Direct expansion} enriches the short request without agentic routing or reference retrieval. \textit{JSON layout} passes the Layout PE specification directly to the generator, including literal text, elements, bounding boxes, typography, and spatial relationships; \textit{Natural} rewrites that plan as prose. \textit{JSON layout} and \textit{Natural} use specifications from the earlier shared-context Layout PE (Appendix~\ref{app:pe}). Human evaluation uses one fixed image per configuration and task.

\paragraph{Shared-prompt generators.}
The generators receive the same enhanced natural-language description for each task, derived from the earlier shared-context Layout PE, without further prompt enhancement. The rewrite preserves requested literal text and expresses coordinates, pixel dimensions, aspect ratios, and colors in prose. Each generator follows its supported resolution and aspect-ratio settings, and one fixed image per task is used for pairwise evaluation.

\subsection{LongTextBench and 7Bench}
\label{app:longtextbench_results}
Baseline scores on LongTextBench are collected from publicly available reports. All 7Bench results reported here are obtained through our own evaluation, as are the results on \llb and \rla.

\paragraph{LongTextBench results.}
Table~\ref{tab:longtextbench} reports the LongTextBench comparison, including the 97.63 overall score of \ours. The listed models score 96.05--98.85 overall, consistent with near-saturation under this coverage metric. We therefore treat LongTextBench as supplementary evidence and use the substantially longer targets of \llb to probe sustained text rendering in the main evaluation.

\begin{table}[htbp]
\centering
\begin{minipage}[t]{0.57\linewidth}
\centering
\small
\caption{Comparison on LongTextBench.}
\label{tab:longtextbench}
\begin{tabular}{lccc}
\toprule
Model & EN $\uparrow$ & ZH $\uparrow$ & Overall $\uparrow$ \\
\midrule
SenseNova-U1.5-8B-MoT & 98.80 & 98.90 & 98.85 \\
Seedream 4.5 & 98.90 & 98.70 & 98.80 \\
HiDream-O1-Image-Pro & 98.20 & 98.00 & 98.10 \\
HiDream-O1-Image & 97.90 & 97.80 & 97.85 \\
\textbf{MOSS-Image} & \textbf{98.08} & \textbf{97.17} & \textbf{97.63} \\
ERNIE-Image + PE & 98.00 & 96.60 & 97.30 \\
Nano Banana 2.0 & 98.00 & 96.50 & 97.25 \\
SenseNova-U1-8B-MoT & 97.90 & 96.20 & 97.05 \\
Qwen-Image-2512 & 95.60 & 96.50 & 96.05 \\
GPT Image 2 & 96.00 & 96.10 & 96.05 \\
\bottomrule
\end{tabular}
\end{minipage}
\end{table}

\paragraph{7Bench tasks and inputs.}
\label{app:7bench_protocol}
7Bench~\citep{izzo20257bench} evaluates layout-guided image generation on 224 prompt--layout pairs, with 32 tasks in each of seven categories: object binding, small bounding boxes, overlapping bounding boxes, color binding, attribute binding, object relationships, and complex compositions. Each task contains a scene description and one to four objects with target bounding boxes. The released annotations specify boxes as $(x_{\min},y_{\min},x_{\max},y_{\max})$. These annotations define the desired object positions and sizes for evaluating spatial control.

\paragraph{7Bench generation and scoring.}
Each run contains one generated image per task, covering 224 images and 576 target-object instances. GPT Image~2, Nano Banana Pro (Gemini~3 Pro Image), Seedream~5.0 Pro, and Qwen Image~3.0 Pro receive the original vision JSON prompts and generate at $1024\times1024$. Ideogram~4 is evaluated with its publicly released FP8 checkpoint, the higher-precision of its two public quantized variants (FP8 and NF4), at CFG~4. Its prompts are converted from the 7Bench annotations into its JSON input schema using a fixed rule applied uniformly to all tasks, preserving the scene description, object phrases, and bounding boxes without manual editing.
We use the Extended TIFA evaluation pipeline with OWL-ViT to localize target objects. For each target box, the scorer selects the same-category detection with the largest intersection-over-union (IoU); a missing detection receives zero. Let $u_j$ denote the resulting IoU for target instance $j$, $N=576$, and $D=\{j:u_j>0\}$. We report
\[
\mathrm{mIoU}=\frac{1}{N}\sum_{j=1}^{N}u_j,
\qquad
\mathrm{mIoU}_{\mathrm{det}}=\frac{1}{|D|}\sum_{j\in D}u_j,
\qquad
\mathrm{DetRate}=\frac{|D|}{N}.
\]
Thus, $\mathrm{mIoU}_{\mathrm{det}}$ measures localization quality for objects with positive detected overlap, while mIoU includes all target instances. Layout AUC is the arithmetic mean of the evaluator's threshold accuracies at IoU thresholds $0.1,0.2,\ldots,0.9$:
\[
\mathrm{LayoutAUC}=\frac{1}{9}\sum_{k=1}^{9}\mathrm{Acc}@\frac{k}{10}.
\]
The threshold accuracies are reported in Table~\ref{tab:7bench_thresholds}.

\paragraph{Filtered outputs and results.}
GPT Image~2 has 22 safety-blocked requests out of 224; their white placeholder images are retained in evaluation and contribute zero IoU. Ideogram~4 also produces filtered outputs. Table~\ref{tab:7bench} reports $\mathrm{mIoU}_{\mathrm{det}}$ alongside Layout AUC, separating localization quality among detected objects from overall threshold accuracy. Table~\ref{tab:7bench_full_results} additionally reports mIoU over all 576 target instances and the positive-overlap detection counts and rates. \ours achieves $\mathrm{mIoU}_{\mathrm{det}}=0.8786$, Layout AUC $=0.7031$, and mIoU $=0.6788$, with positive overlap for 445 of 576 target instances. The Ideogram~4 report lists a 7Bench mIoU of about 0.69~\citep{ideogram-4-2026}, whereas we measure 0.2433 (0.4449 over detected instances). Our evaluation uses the public FP8 checkpoint, rule-converted prompts, and the Extended TIFA pipeline, with filtered outputs and missing detections scored as zero. Differences in model variant, prompt format, and evaluation implementation may account for the gap.

\begin{table}[htbp]
\centering
\small
\caption{Detailed layout-control results on 7Bench. All runs contain 224 images and 576 target objects. Detected objects have IoU greater than zero.}
\label{tab:7bench_full_results}
\resizebox{\linewidth}{!}{
\begin{tabular}{lccrcc}
\toprule
Model & mIoU $\uparrow$ & $\mathrm{mIoU}_{\mathrm{det}}$ $\uparrow$ & Detected & Det. rate $\uparrow$ & Layout AUC $\uparrow$ \\
\midrule
GPT Image 2 & 0.1431 & 0.3042 & 271 & 0.4705 & 0.1554 \\
Nano Banana Pro & 0.1493 & 0.3296 & 261 & 0.4531 & 0.1628 \\
Qwen Image 3.0 Pro & 0.1694 & 0.3148 & 310 & 0.5382 & 0.1758 \\
Seedream 5.0 Pro & 0.1670 & 0.3273 & 294 & 0.5104 & 0.1794 \\
Ideogram 4.0 & 0.2433 & 0.4449 & 315 & 0.5469 & 0.2046 \\
\ours & \textbf{0.6788} & \textbf{0.8786} & \textbf{445} & \textbf{0.7726} & \textbf{0.7031} \\
\bottomrule
\end{tabular}
}
\end{table}

\begin{table}[htbp]
\centering
\scriptsize
\caption{7Bench accuracy at IoU thresholds from 0.1 to 0.9. Layout AUC is the arithmetic mean of these nine scores.}
\label{tab:7bench_thresholds}
\resizebox{\linewidth}{!}{
\begin{tabular}{lccccccccc}
\toprule
Model & Acc@0.1 & Acc@0.2 & Acc@0.3 & Acc@0.4 & Acc@0.5 & Acc@0.6 & Acc@0.7 & Acc@0.8 & Acc@0.9 \\
\midrule
GPT Image 2 & 0.4055 & 0.3337 & 0.2608 & 0.1749 & 0.1194 & 0.0662 & 0.0290 & 0.0089 & 0.0000 \\
Nano Banana Pro & 0.4018 & 0.3441 & 0.2727 & 0.2046 & 0.1287 & 0.0748 & 0.0320 & 0.0067 & 0.0000 \\
Qwen Image 3.0 Pro & 0.4647 & 0.3925 & 0.2980 & 0.2083 & 0.1339 & 0.0614 & 0.0216 & 0.0015 & 0.0000 \\
Seedream 5.0 Pro & 0.4803 & 0.3973 & 0.3114 & 0.2087 & 0.1194 & 0.0714 & 0.0246 & 0.0011 & 0.0000 \\
Ideogram 4.0 & 0.4301 & 0.3813 & 0.3311 & 0.2786 & 0.2161 & 0.1484 & 0.0487 & 0.0052 & 0.0015 \\
\ours & \textbf{0.7556} & \textbf{0.7511} & \textbf{0.7481} & \textbf{0.7336} & \textbf{0.7295} & \textbf{0.7284} & \textbf{0.7221} & \textbf{0.6942} & \textbf{0.4650} \\
\bottomrule
\end{tabular}
}
\end{table}

\clearpage
\section{Supplementary Rubric-Based Evaluation}
\label{app:rubric_evaluation}
We additionally evaluate the 61 \rla tasks with a task-specific GPT-6 scoring rubric. This evaluation provides dimension-level diagnostics alongside the human-preference results in the main paper.

\paragraph{Task-specific criteria.}
The task set contains 20 posters, 20 infographics, and 21 user interfaces. Each rubric is derived from the original short request. All outputs, including those generated from detailed or enhanced prompts, are judged against the same request. Layout criteria reflect the requested genre or the typical organization of a named product. Unspecified copy, colors, and composition remain open to design choices; additional content or visual complexity is rewarded only when it serves the request.

\paragraph{Dimensions and aggregation.}
The six dimensions are theme and explicit requirements (including obvious factual errors), layout and genre correctness, text readability, content coherence, aesthetics and visual hierarchy, and rendering integrity. Their weights are $\mathbf{w}=(30,20,15,10,20,5)$. Each dimension is scored from 0 to 5: 0 denotes missing or unusable core requirements, 1 severe problems, 2 partial fulfillment, 3 the main requirements with visible problems, 4 minor local defects, and 5 clear and coordinated fulfillment. For model $m$ on task $i$, we compute
\[
s_{m,i}=\sum_{d=1}^{6}w_d\frac{r_{m,i,d}}{5},
\qquad S_m=\frac{1}{61}\sum_{i=1}^{61}s_{m,i}.
\]
Thus $S_m$ lies in $[0,100]$ and weights tasks equally; equal totals are tied. Failed, blank, garbled, or blocked outputs remain in the denominator, and a blank image without an identifiable design receives six zeros. Output-size compliance is recorded separately and does not enter the visual score. Checks within a dimension are considered jointly rather than weighted by their count, and defects are not indiscriminately penalized across multiple dimensions.

\paragraph{Judging procedure.}
Each image receives one primary GPT-6 assessment using the anonymized image, the original short request, and its task-specific rubric. Model identities and human-preference rankings are hidden from the judge. Targeted visual spot checks identify scoring issues. Factual content is classified as showing no obvious error, containing a concrete error, or being unverifiable. Identified factual errors affect the theme-and-requirements score, without an additional penalty to the total. Plausible fictional details are permitted when appropriate to the task.

\paragraph{Configurations and results.}
Table~\ref{tab:rubric_results} reports separate GPT-6 rubric evaluations of end-to-end systems, prompt strategies, and generators under enhanced natural-language prompts. In the system comparison, \ours obtains 94.93, compared with 97.72 for GPT Image~2. Under shared enhanced natural-language prompts, \ours obtains 87.62, followed by SenseNova U1 at 79.51. The same \ours natural-language outputs are used in both ablations. The \ours JSON-layout entries in panels (a) and (b) use the final and earlier Layout PE versions, respectively (Appendix~\ref{app:pe}), and are therefore not directly comparable.

\paragraph{Interpretation.}
The rubric provides dimension-level diagnostics but limited separation among the strongest configurations. In the prompt-enhancement ablation, enhancement without agentic planning, detailed prompts, and JSON layout score 95.59, 95.46, and 95.07, respectively, a spread of only 0.52 points. This requirement-focused scale is less sensitive to fine differences in complex content organization and visual design. We therefore use human preferences as the primary ranking evidence and the rubric as supplementary analysis. Because GPT-6 serves as both planner and automated judge, the human evaluation also provides an independent assessment of the resulting designs.

\begin{table}[p]
\centering
\footnotesize
\setlength{\tabcolsep}{2pt}
\renewcommand{\arraystretch}{1.04}
\caption{Supplementary GPT-6 rubric-based evaluation on the 61 \rla tasks. Each image is judged against the original short request and its task-specific rubric. Dimension means are on a 0--5 scale; Total is the weighted score on a 0--100 scale. Bold totals mark the best score within each panel. These automatic scores complement the human-preference results.}
\label{tab:rubric_results}
\begin{minipage}{\linewidth}
\textbf{(a) End-to-end systems}\par\smallskip
\begin{tabular*}{\linewidth}{@{\extracolsep{\fill}}lrrrrrrr@{}}
\toprule
Model & Topic & Layout & Text & Coh. & Aesth. & Rend. & Total $\uparrow$ \\
\midrule
GPT Image 2 & 4.77 & 4.92 & 4.95 & 4.79 & 5.00 & 5.00 & \textbf{97.72} \\
\textbf{MOSS-Image + Layout PE} & 4.72 & 4.80 & 4.62 & 4.66 & 4.84 & 4.87 & 94.93 \\
Qwen Image 3.0 Pro + PE & 4.48 & 4.82 & 4.61 & 4.59 & 4.89 & 4.97 & 93.64 \\
Seedream 5.0 Pro & 4.39 & 4.56 & 4.11 & 4.11 & 4.62 & 4.87 & 88.52 \\
Nano Banana Pro & 4.23 & 4.41 & 3.92 & 3.85 & 4.46 & 4.84 & 85.15 \\
ERNIE Image + PE & 3.52 & 3.80 & 4.56 & 3.90 & 4.10 & 4.92 & 79.15 \\
Hunyuan Image 3 Instruct + PE & 3.84 & 3.82 & 4.11 & 3.66 & 4.05 & 4.80 & 78.95 \\
Boogu Image + PE & 3.70 & 3.92 & 3.62 & 3.62 & 3.77 & 4.56 & 75.66 \\
Ideogram 4 + local PE & 2.43 & 2.25 & 1.90 & 2.16 & 2.13 & 3.46 & 45.56 \\
\bottomrule
\end{tabular*}

\end{minipage}
\par\vspace{0.8em}
\begin{minipage}{\linewidth}
\textbf{(b) Prompt strategies on a fixed MOSS-Image checkpoint}\par\smallskip
\begin{tabular*}{\linewidth}{@{\extracolsep{\fill}}lrrrrrrr@{}}
\toprule
Prompt strategy & Topic & Layout & Text & Coh. & Aesth. & Rend. & Total $\uparrow$ \\
\midrule
Enhanced without agentic & 4.79 & 4.85 & 4.77 & 4.67 & 4.75 & 4.79 & \textbf{95.59} \\
Detailed & 4.84 & 4.87 & 4.72 & 4.61 & 4.70 & 4.77 & 95.46 \\
Enhanced JSON layout & 4.82 & 4.87 & 4.61 & 4.64 & 4.72 & 4.69 & 95.07 \\
Enhanced natural & 4.59 & 4.43 & 4.21 & 4.03 & 4.31 & 4.43 & 87.62 \\
Short & 2.33 & 2.67 & 1.30 & 1.64 & 2.74 & 3.95 & 46.72 \\
\bottomrule
\end{tabular*}

\end{minipage}
\par\vspace{0.8em}
\begin{minipage}{\linewidth}
\textbf{(c) Generators under shared enhanced-natural prompts}\par\smallskip
\begin{tabular*}{\linewidth}{@{\extracolsep{\fill}}lrrrrrrr@{}}
\toprule
Model & Topic & Layout & Text & Coh. & Aesth. & Rend. & Total $\uparrow$ \\
\midrule
\textbf{MOSS-Image} & 4.59 & 4.43 & 4.21 & 4.03 & 4.31 & 4.43 & \textbf{87.62} \\
SenseNova U1 & 4.20 & 4.16 & 3.49 & 3.51 & 3.97 & 4.31 & 79.51 \\
ERNIE Image & 3.77 & 3.26 & 3.02 & 2.72 & 3.46 & 4.16 & 68.16 \\
Ideogram 4 & 3.70 & 3.11 & 2.95 & 2.79 & 2.74 & 3.80 & 63.87 \\
Qwen Image 2512 & 2.80 & 2.89 & 1.67 & 1.97 & 3.46 & 4.16 & 55.31 \\
Boogu Image & 3.13 & 2.95 & 1.64 & 2.07 & 2.95 & 3.64 & 55.08 \\
FLUX.2 & 2.07 & 2.59 & 0.89 & 1.51 & 3.18 & 3.95 & 45.10 \\
Krea 2 Raw & 2.39 & 2.66 & 0.97 & 1.70 & 2.46 & 3.31 & 44.44 \\
Z-Image Base & 2.49 & 2.33 & 1.16 & 1.56 & 2.30 & 2.74 & 42.79 \\
Hunyuan Image 3 Instruct & 2.05 & 2.26 & 1.13 & 1.57 & 2.39 & 3.28 & 40.74 \\
LongCat Image & 1.72 & 2.18 & 0.80 & 1.25 & 2.66 & 3.33 & 37.90 \\
HiDream-O1 (dev-2604) & 1.26 & 1.15 & 1.64 & 1.39 & 1.36 & 2.52 & 27.84 \\
\bottomrule
\end{tabular*}

\end{minipage}
\par\smallskip
\begin{minipage}{\linewidth}
\footnotesize
Topic: theme and explicit requirements; Coh.: content coherence; Aesth.: aesthetics; Rend.: rendering integrity. Weights are 30\%, 20\%, 15\%, 10\%, 20\%, and 5\%, respectively. Local PE denotes a Qwen3VL-8B-based prompt rewriter. The same \ours enhanced-natural images and scores are used in panels (b) and (c). The \ours JSON-layout entries in panels (a) and (b) use different Layout PE versions (Appendix~\ref{app:pe}).
\end{minipage}
\end{table}

\clearpage
\section{Layout-Guided Control: Case Studies}
\label{app:layout_editing_cases}
\begin{figure}[htbp]
\centering
\begin{minipage}{0.49\linewidth}
\centering\small\textbf{(a) Reorder steps 2 and 3}
\end{minipage}\hfill
\begin{minipage}{0.49\linewidth}
\centering\small\textbf{(b) Replace the main title}
\end{minipage}
\par\smallskip
\begin{minipage}[t]{0.235\linewidth}
\centering
{\small\textbf{Before}\par}\smallskip
\includegraphics[width=\linewidth,height=3.7cm,keepaspectratio]{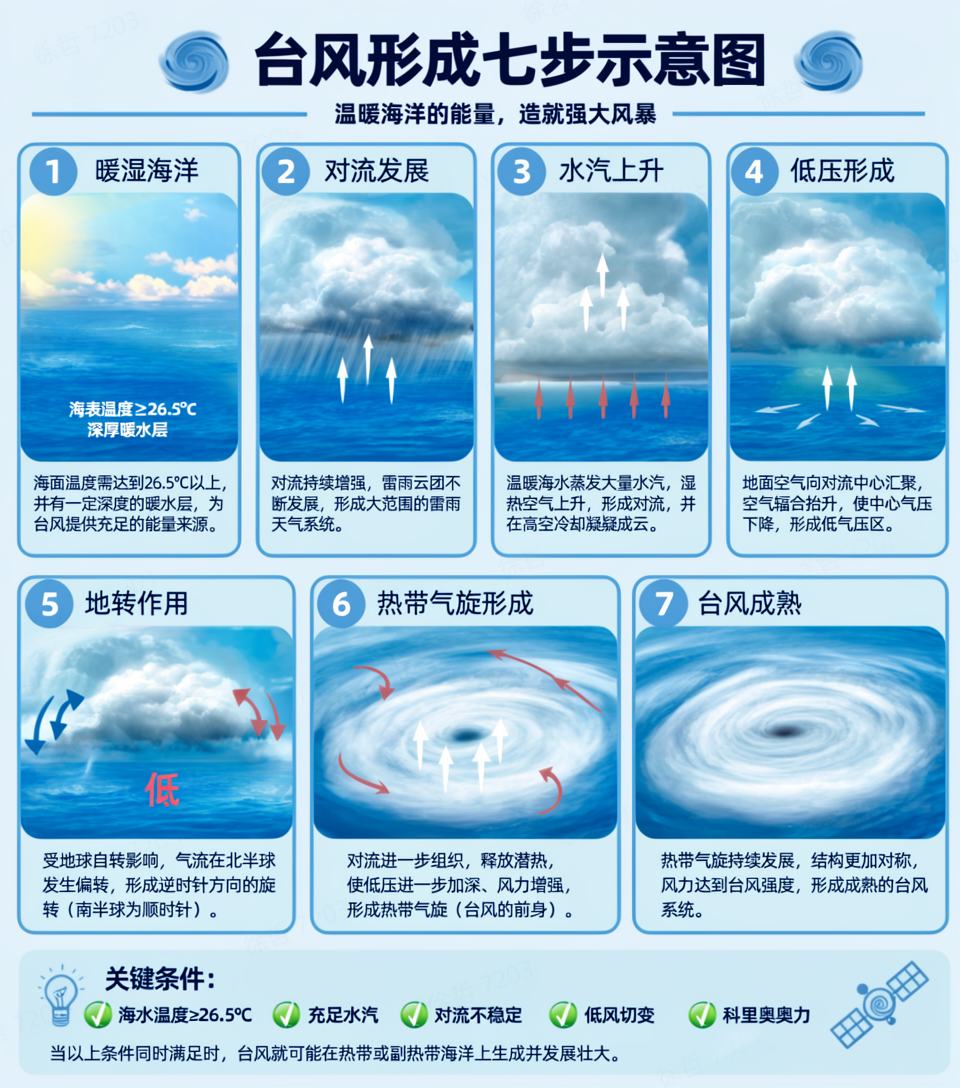}
\end{minipage}\hfill
\begin{minipage}[t]{0.235\linewidth}
\centering
{\small\textbf{After}\par}\smallskip
\includegraphics[width=\linewidth,height=3.7cm,keepaspectratio]{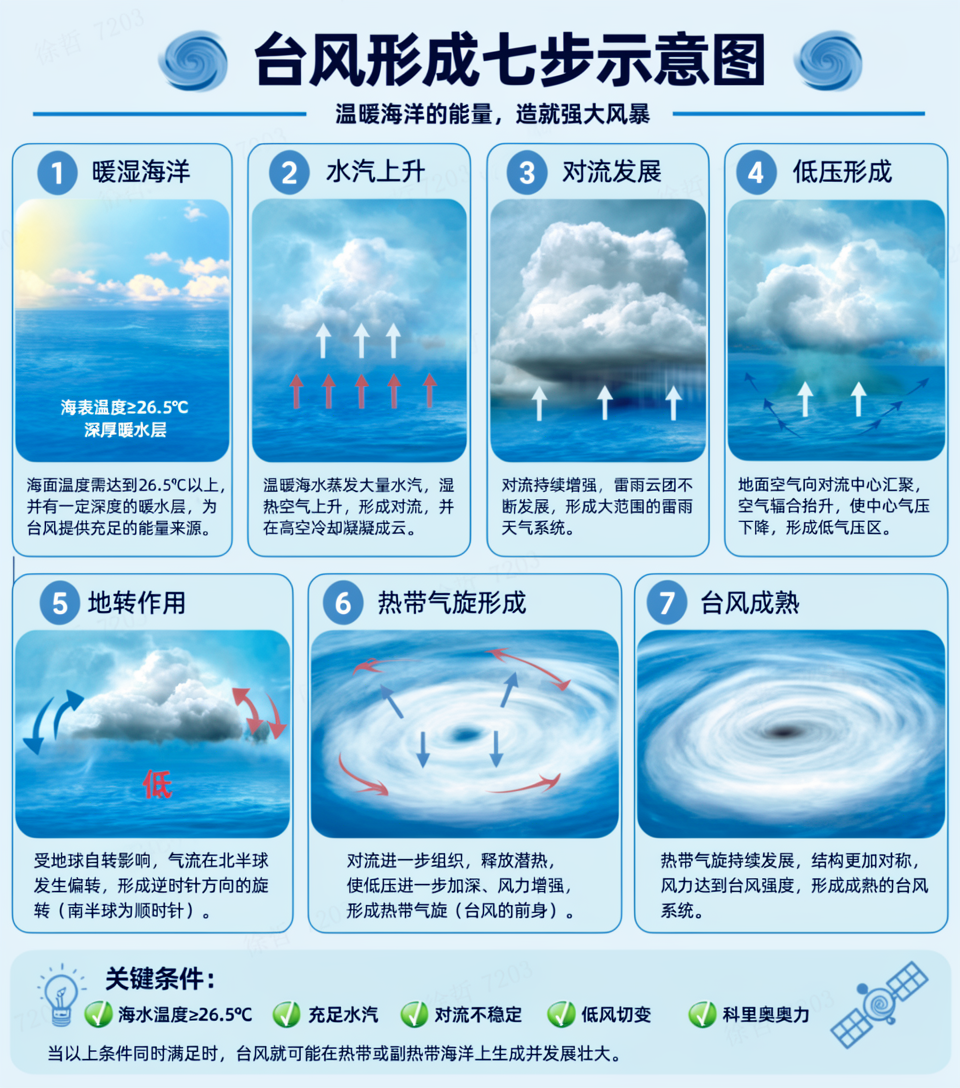}
\end{minipage}\hfill
\begin{minipage}[t]{0.235\linewidth}
\centering
{\small\textbf{Before}\par}\smallskip
\includegraphics[width=\linewidth,height=3.7cm,keepaspectratio]{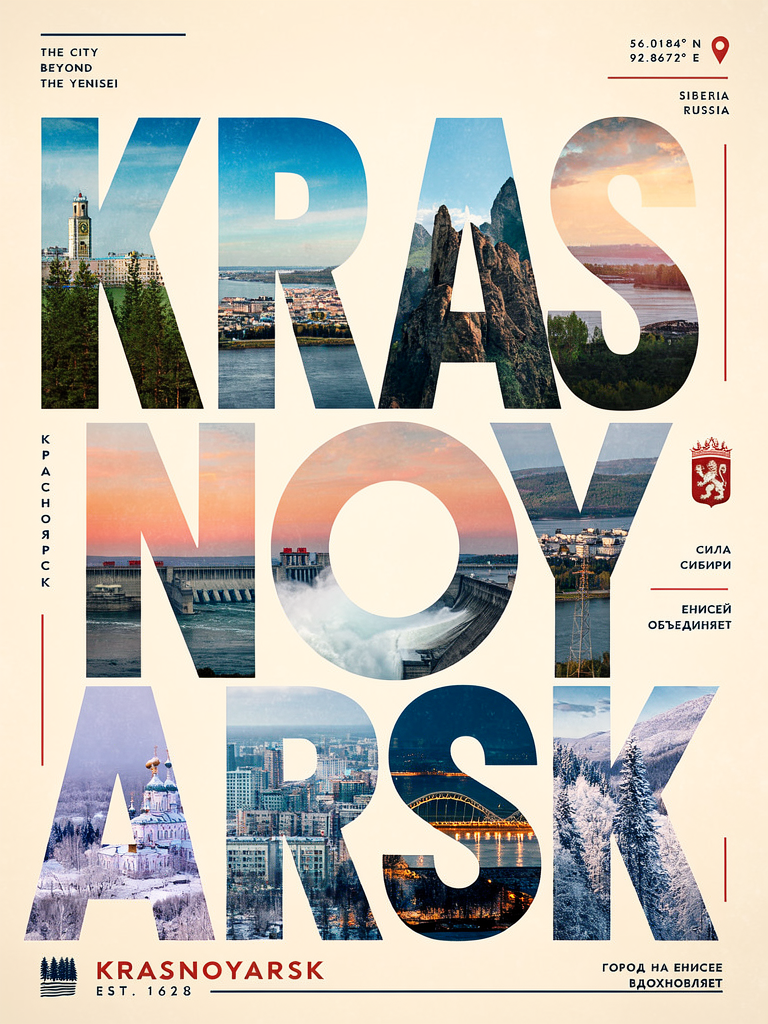}
\end{minipage}\hfill
\begin{minipage}[t]{0.235\linewidth}
\centering
{\small\textbf{After}\par}\smallskip
\includegraphics[width=\linewidth,height=3.7cm,keepaspectratio]{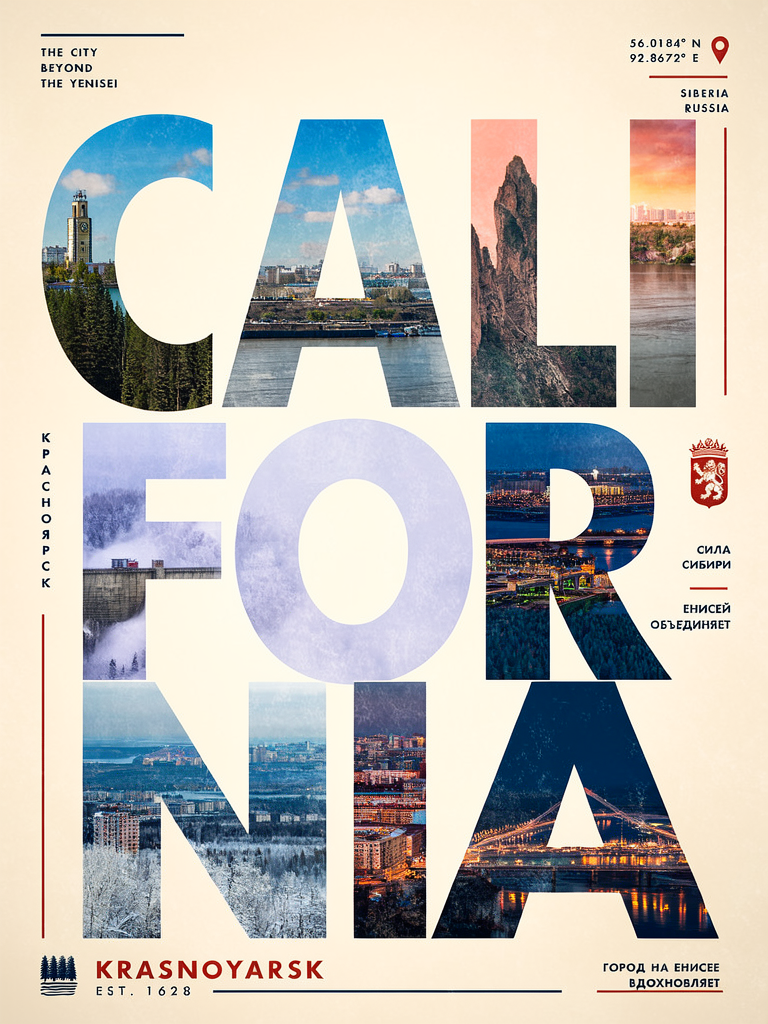}
\end{minipage}
\caption{\textbf{Qualitative editing cases under layout constraints.} (a) Reordering the second and third infographic steps together with their associated content. (b) Replacing \texttt{KRASNOYARSK} with \texttt{CALIFORNIA} while retaining the photograph-filled lettering style and overall composition. Within each pair, left: before; right: after. Local visual details vary.}
\label{fig:layout_editing_cases}
\end{figure}

Figure~\ref{fig:layout_editing_cases} presents two qualitative editing cases: rearranging information blocks and replacing prominent text while retaining the overall design.

\paragraph{Reordering information blocks.}
In the typhoon infographic, the second and third panels exchange positions: water-vapor ascent precedes convection development in the revised version. The associated headings, explanatory text, and illustrations follow the new order, while the seven-panel structure, blue palette, title banner, and bottom summary remain recognizable. The change therefore involves coordinating several related elements within a common layout.

\paragraph{Replacing prominent text.}
In the city poster, the dominant three-line title changes from \texttt{KRASNOYARSK} to \texttt{CALIFORNIA}. The revised design retains the large photograph-filled letters, light background, red accents, and peripheral annotations. Letter shapes and interior imagery adapt to the new word while preserving the typographic composition. Smaller city labels remain from the original design. Together, the two cases demonstrate targeted changes to content and spatial arrangement while preserving the overall visual style.

\clearpage
\section{Line-Break Alignment: Denoising Case Study}
\label{app:linebreak_cases}
\begin{figure}[htbp]
\centering
\begin{minipage}{\linewidth}
\small\textbf{With line breaks}
\end{minipage}
\par\smallskip
\begin{minipage}[t]{0.19\linewidth}
\centering
{\small Step 1\par}{\footnotesize $\sigma=0.9886$\par}\smallskip
\includegraphics[width=\linewidth]{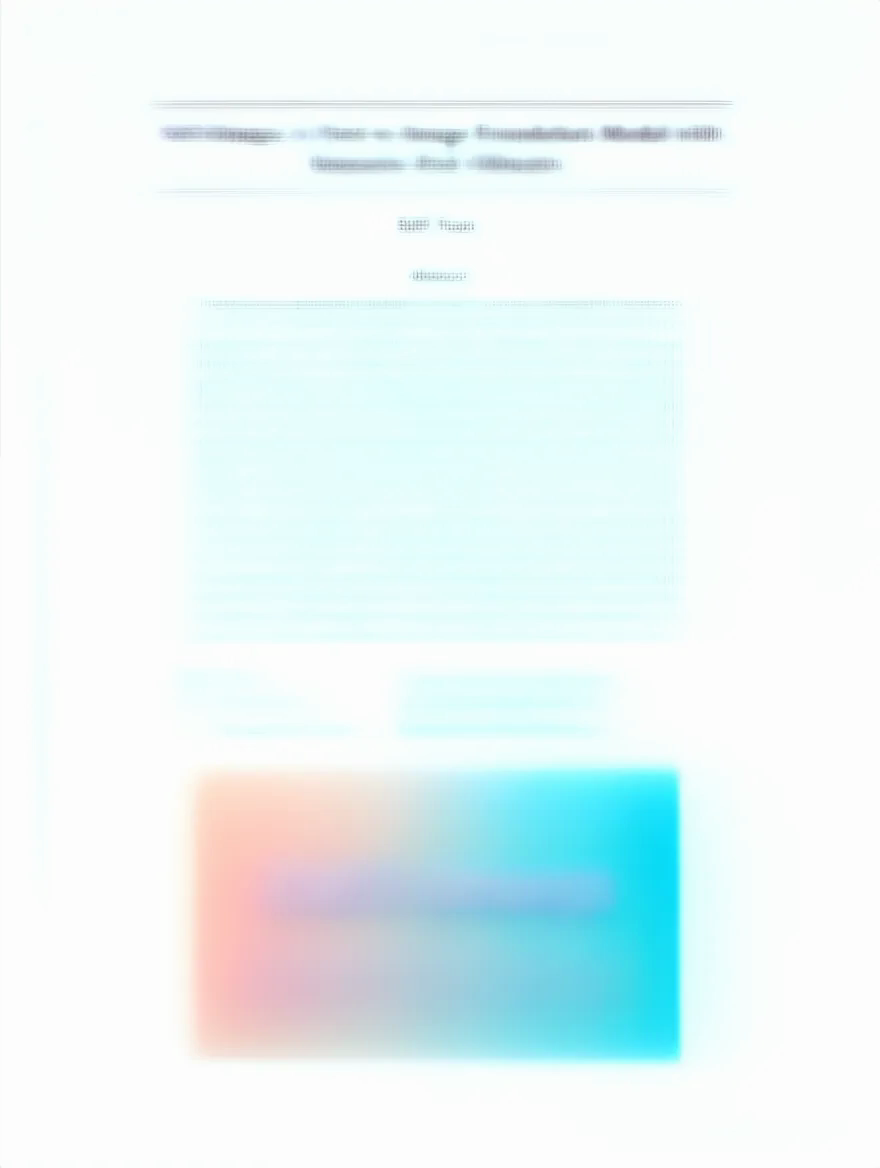}
\end{minipage}\hfill
\begin{minipage}[t]{0.19\linewidth}
\centering
{\small Step 8\par}{\footnotesize $\sigma=0.8919$\par}\smallskip
\includegraphics[width=\linewidth]{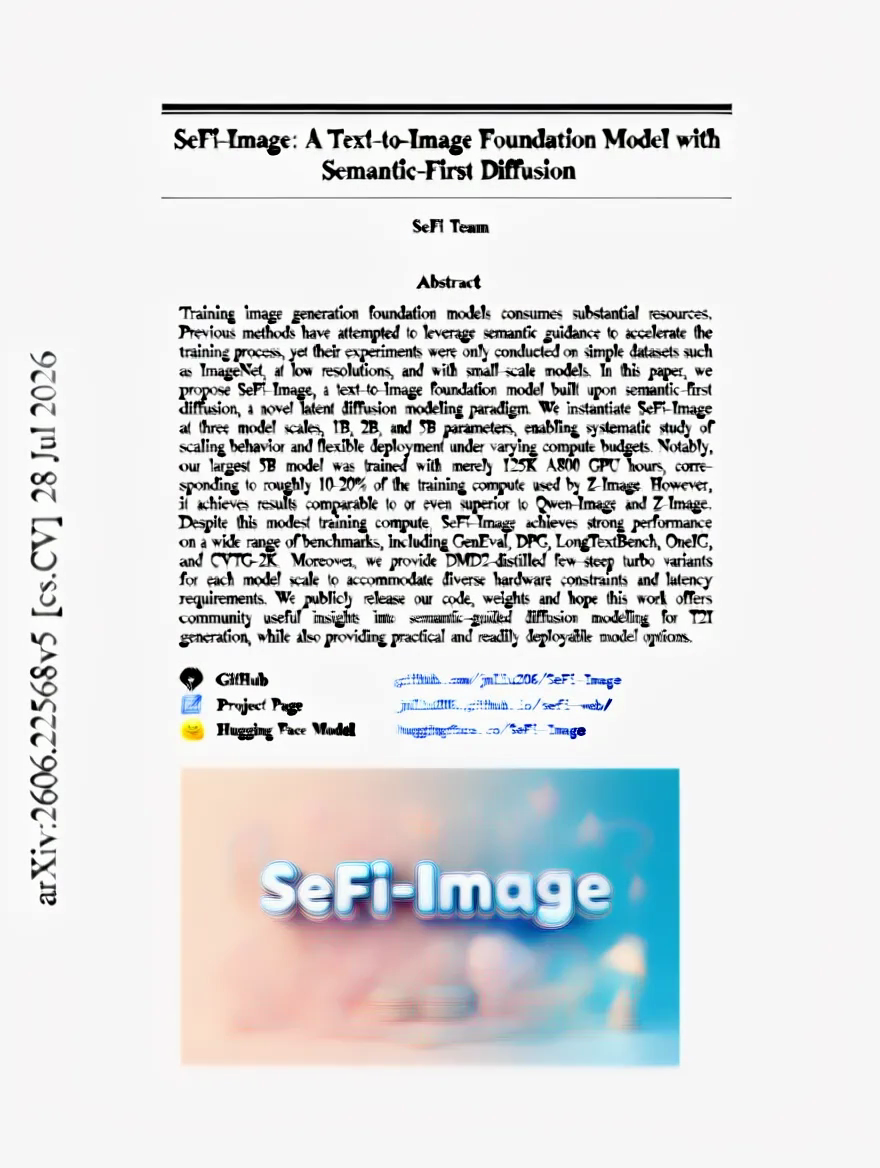}
\end{minipage}\hfill
\begin{minipage}[t]{0.19\linewidth}
\centering
{\small Step 16\par}{\footnotesize $\sigma=0.7241$\par}\smallskip
\includegraphics[width=\linewidth]{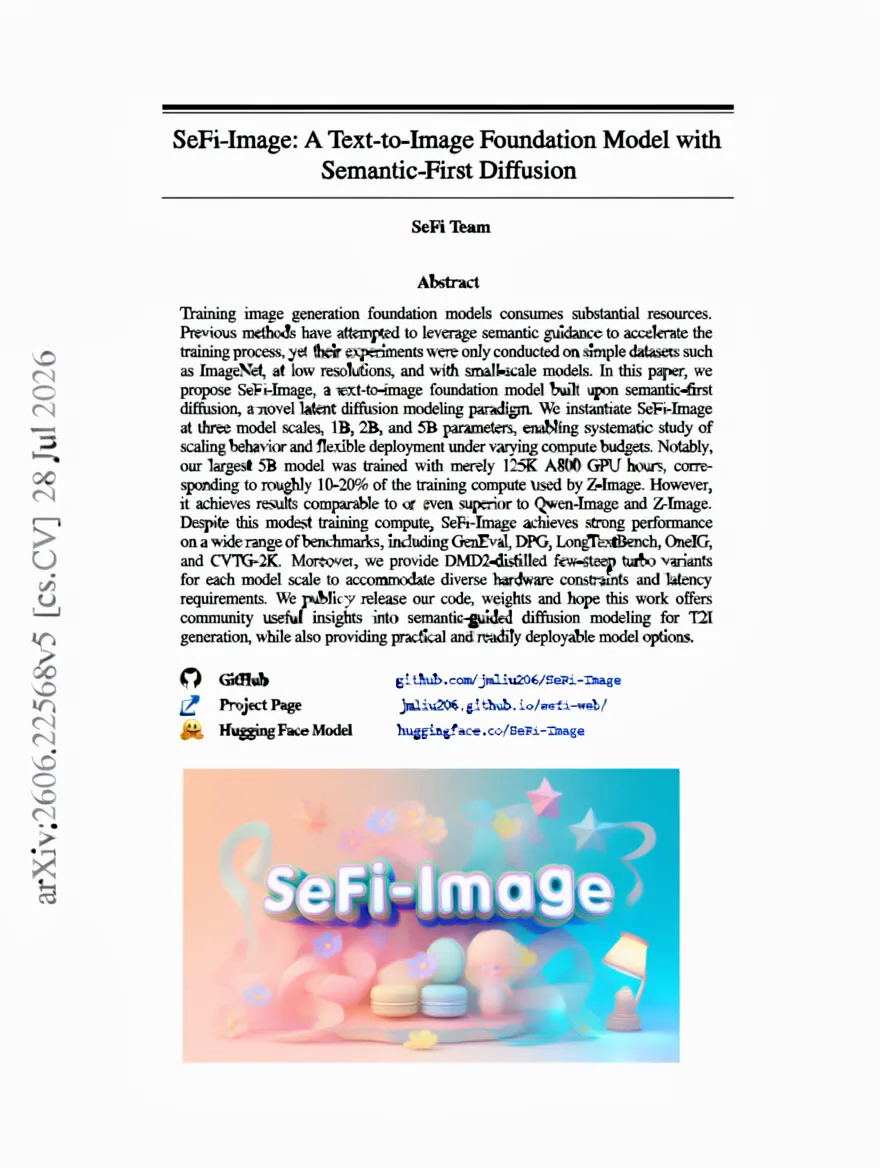}
\end{minipage}\hfill
\begin{minipage}[t]{0.19\linewidth}
\centering
{\small Step 24\par}{\footnotesize $\sigma=0.4286$\par}\smallskip
\includegraphics[width=\linewidth]{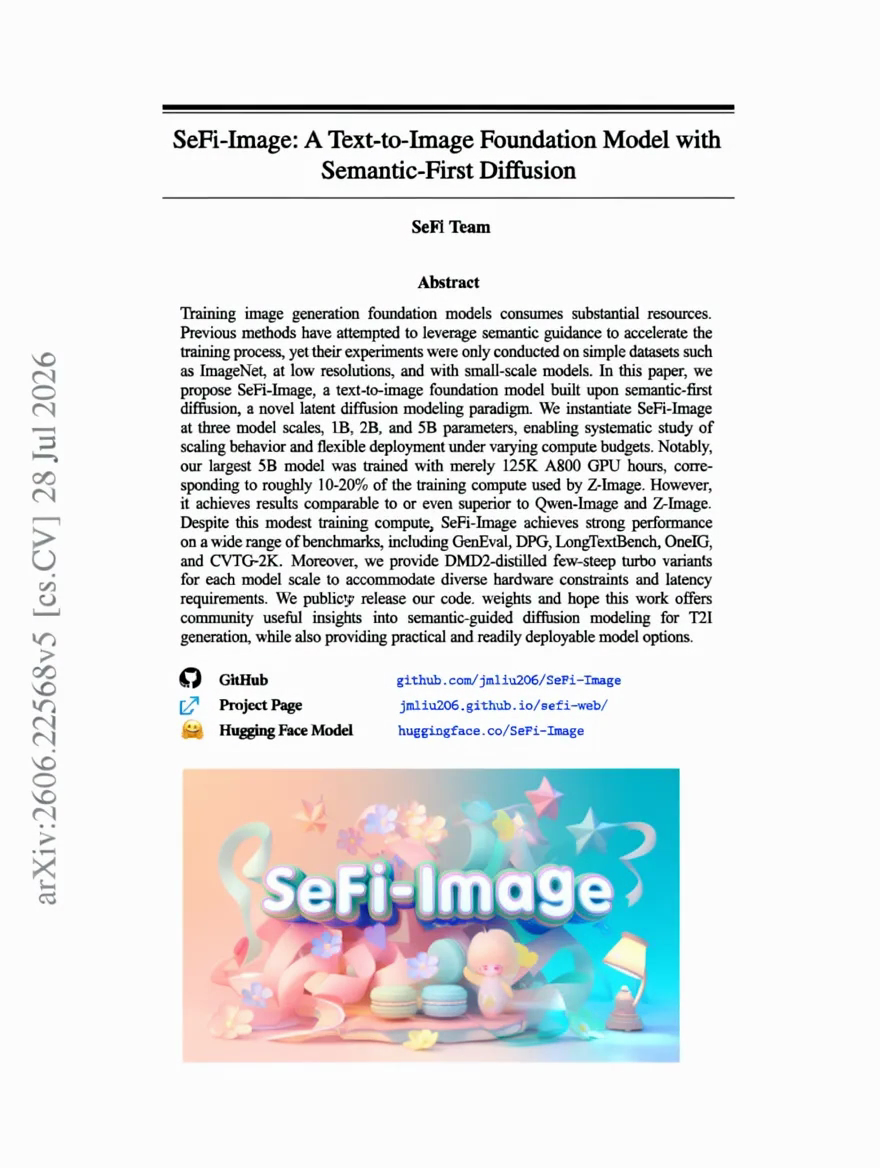}
\end{minipage}\hfill
\begin{minipage}[t]{0.19\linewidth}
\centering
{\small Step 30\par}{\footnotesize $\sigma=0$\par}\smallskip
\includegraphics[width=\linewidth]{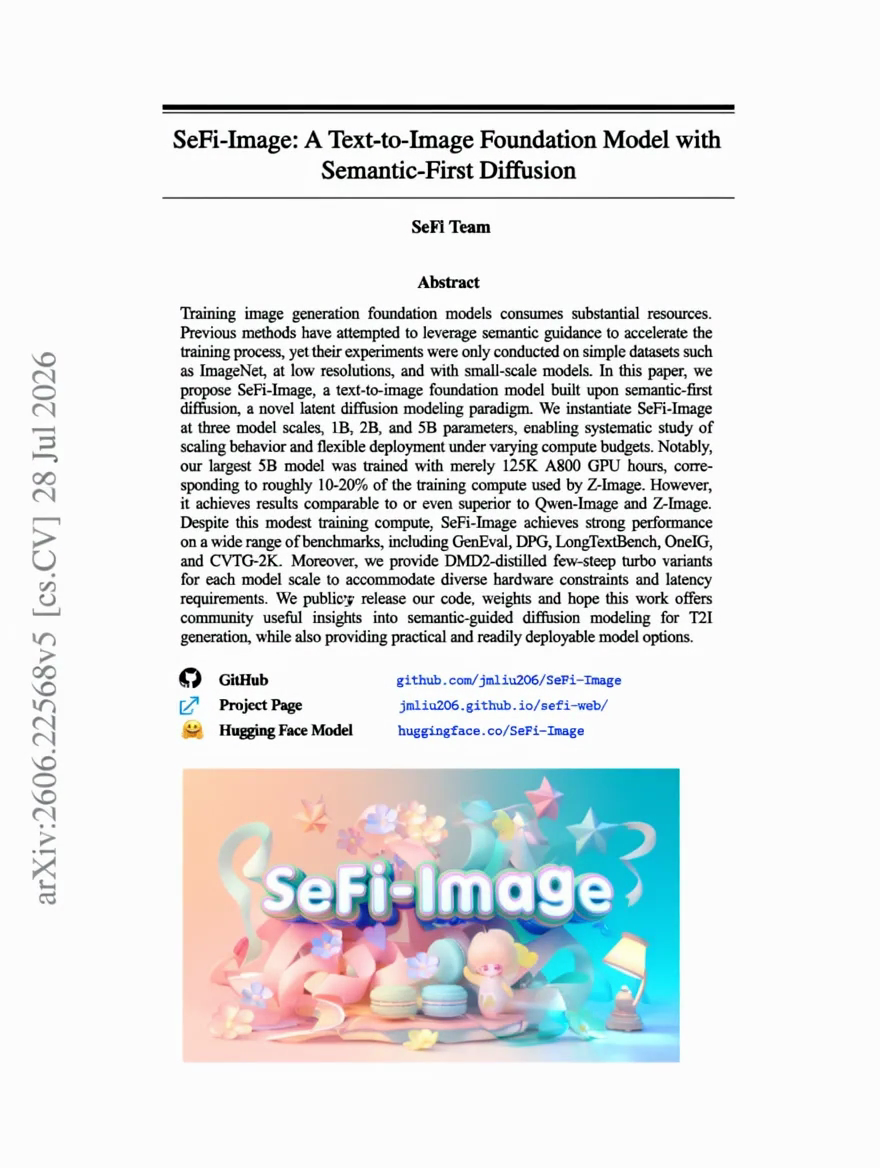}
\end{minipage}
\par\medskip
\begin{minipage}{\linewidth}
\small\textbf{Without line breaks}
\end{minipage}
\par\smallskip
\begin{minipage}[t]{0.19\linewidth}
\centering
{\small Step 1\par}{\footnotesize $\sigma=0.9886$\par}\smallskip
\includegraphics[width=\linewidth]{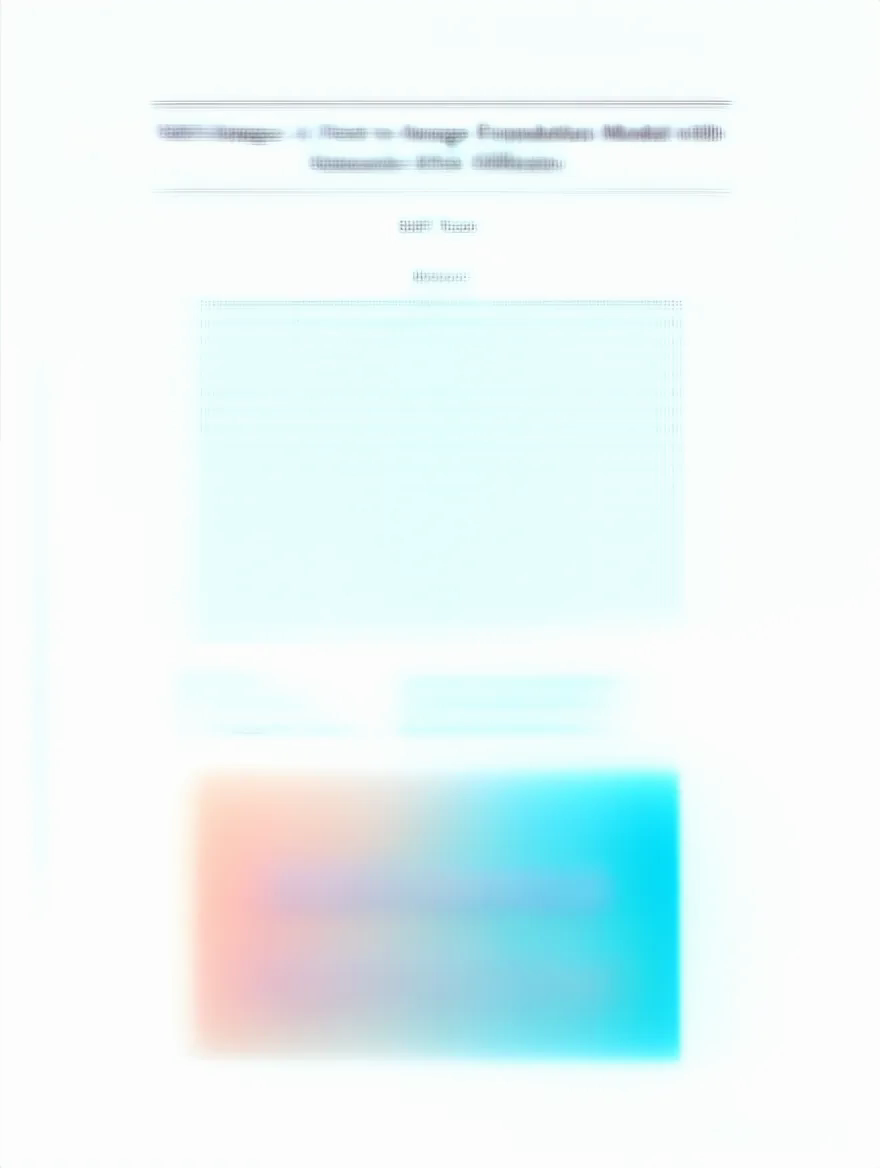}
\end{minipage}\hfill
\begin{minipage}[t]{0.19\linewidth}
\centering
{\small Step 8\par}{\footnotesize $\sigma=0.8919$\par}\smallskip
\includegraphics[width=\linewidth]{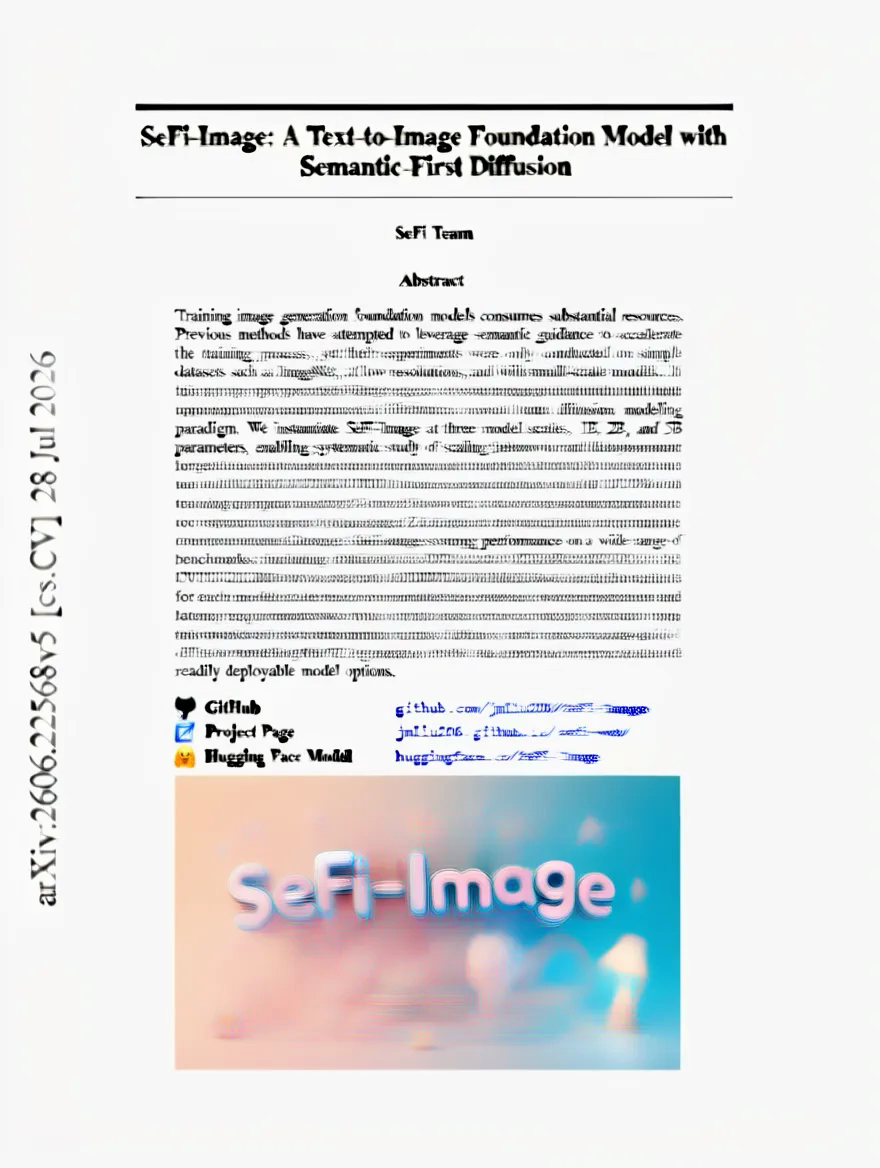}
\end{minipage}\hfill
\begin{minipage}[t]{0.19\linewidth}
\centering
{\small Step 16\par}{\footnotesize $\sigma=0.7241$\par}\smallskip
\includegraphics[width=\linewidth]{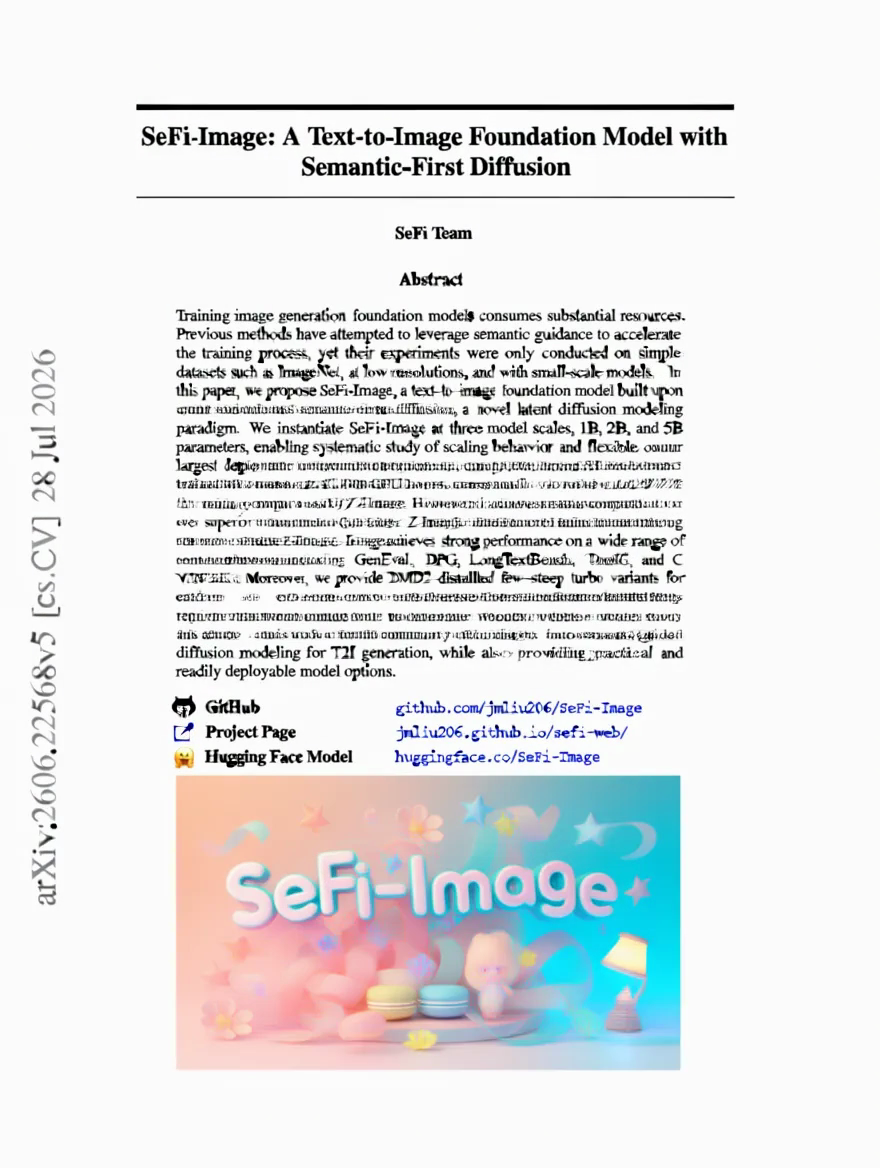}
\end{minipage}\hfill
\begin{minipage}[t]{0.19\linewidth}
\centering
{\small Step 24\par}{\footnotesize $\sigma=0.4286$\par}\smallskip
\includegraphics[width=\linewidth]{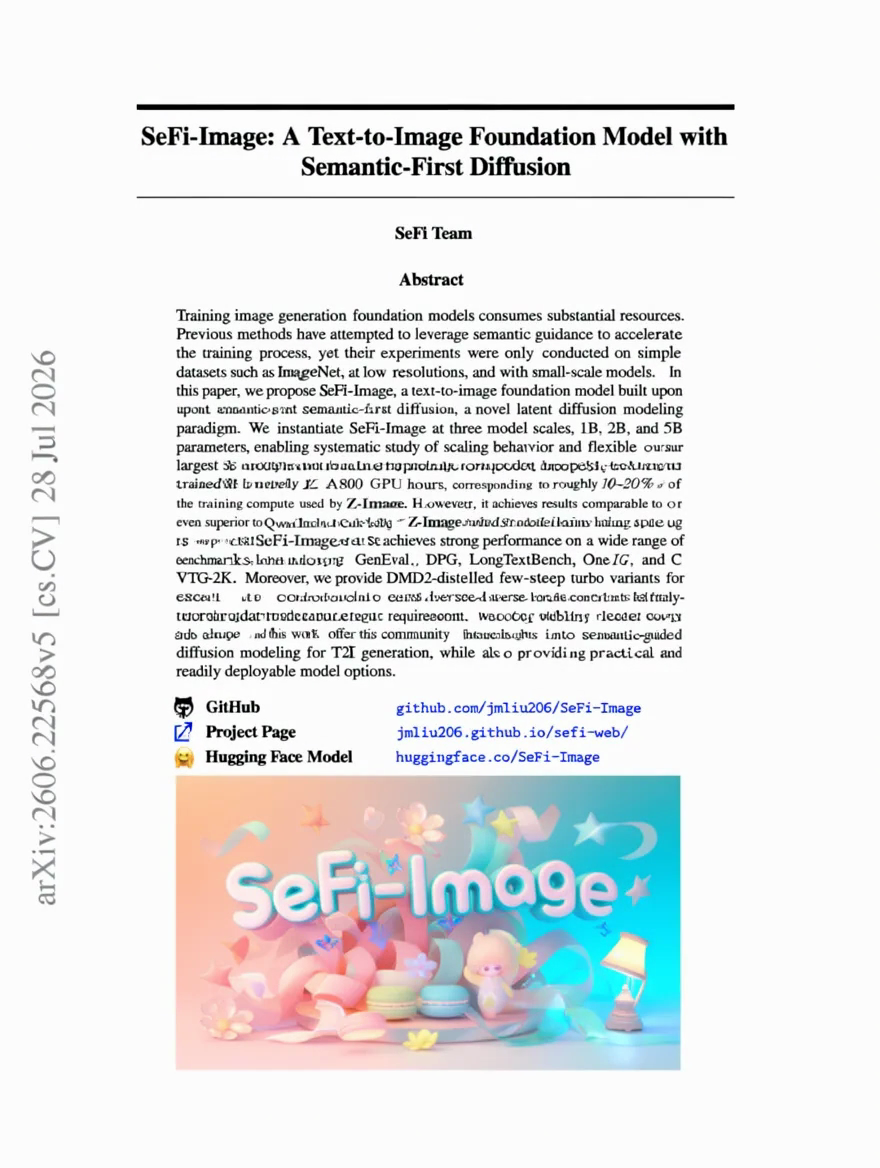}
\end{minipage}\hfill
\begin{minipage}[t]{0.19\linewidth}
\centering
{\small Step 30\par}{\footnotesize $\sigma=0$\par}\smallskip
\includegraphics[width=\linewidth]{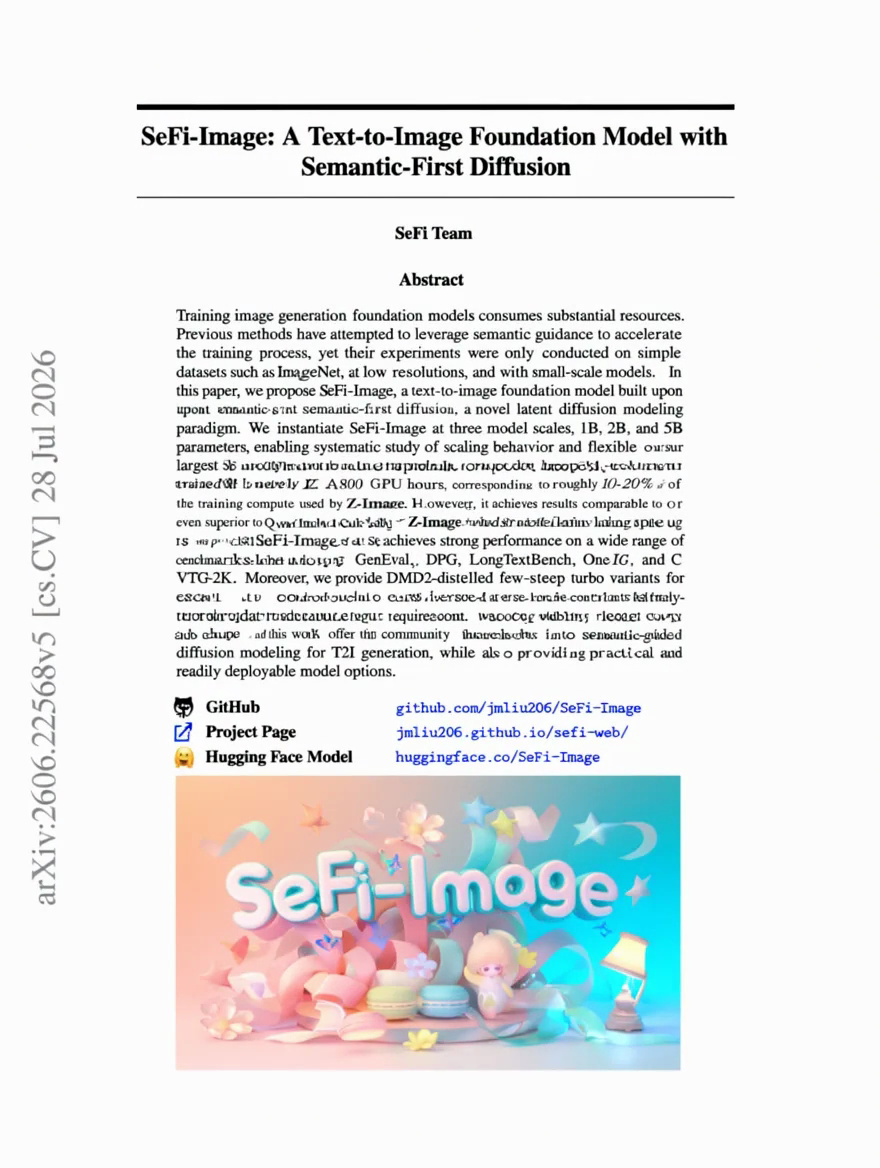}
\end{minipage}
\caption{\textbf{Line-break conditioning during denoising.} Top: prompt with explicit line breaks. Bottom: prompt without line breaks. Each column shows the same denoising step, progressing from left to right; labels give the step number and post-update noise level $\sigma$. Explicit line breaks help the character content of each line become recognizable earlier, with clearer text in the final displayed step.}
\label{fig:linebreak_denoising}
\end{figure}

Figure~\ref{fig:linebreak_denoising} compares the denoising process with and without explicit line breaks in the prompt. The columns show outputs after sampling steps 1, 8, 16, 24, and 30. Both runs use 30 denoising updates with a timestep shift of $s=3$. Starting from the uniform unshifted grid $u_k=1-k/30$, the shifted noise level is
\[
\sigma_k=\frac{s u_k}{1+(s-1)u_k}=\frac{3u_k}{1+2u_k},\qquad k=0,\ldots,30.
\]
The displayed snapshots are recorded after update $k$ and are labeled with the corresponding $\sigma_k$; the final step has $\sigma_{30}=0$.

With explicit line breaks, the character content of many lines is already distinguishable at step 8 and becomes clearer as denoising proceeds. Without line breaks, several rows remain ambiguous at intermediate steps, and malformed characters persist in the final displayed step. This qualitative comparison illustrates how specifying line boundaries can help establish the content of each line earlier and improve the final text rendering.

\clearpage
\section{Bounding-Box Coordinates: Generalization to Unseen Resolutions}
\label{app:bbox_coordinate_cases}
\begin{figure}[htbp]
\centering
\begin{minipage}[t]{0.32\linewidth}
\centering
{\small\textbf{(a) Lower resolution}\par Comparison\par}\smallskip
\includegraphics[width=\linewidth,height=3.2cm,keepaspectratio]{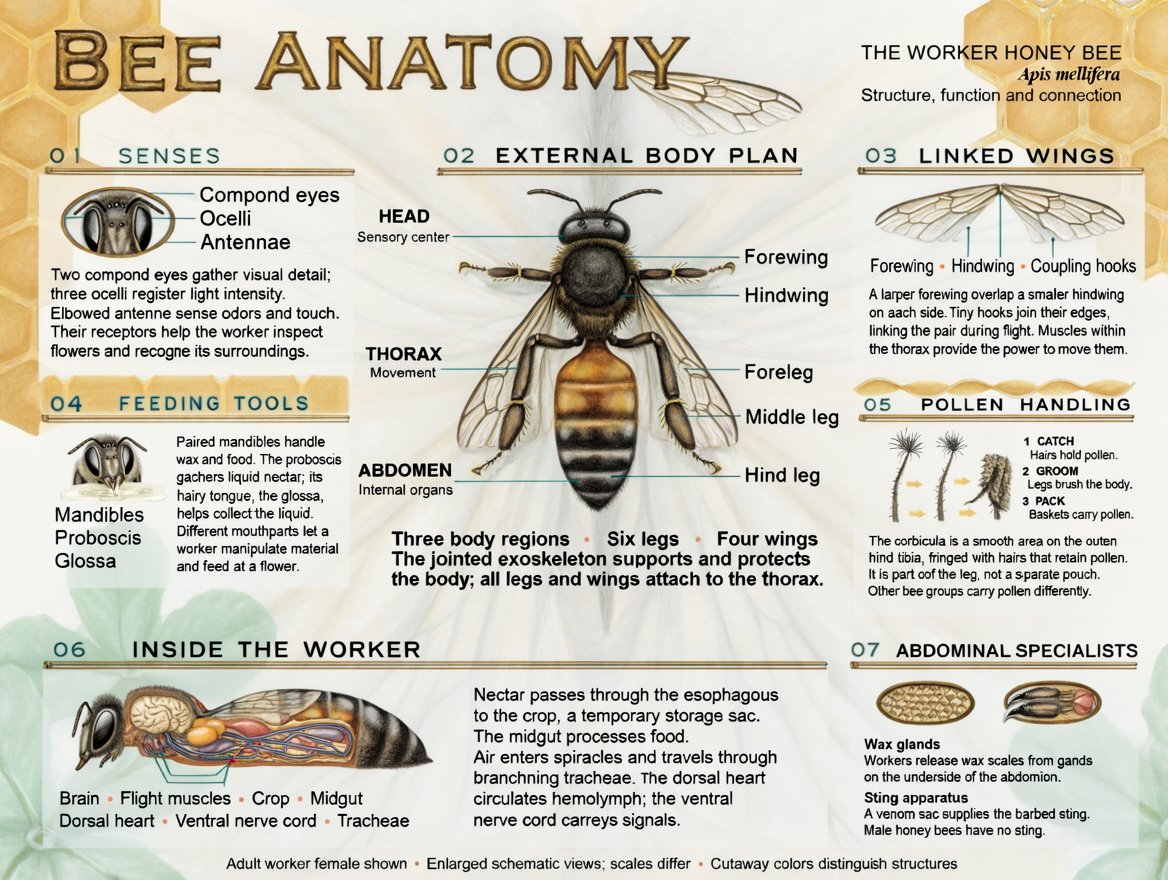}
\end{minipage}\hfill
\begin{minipage}[t]{0.32\linewidth}
\centering
{\small\textbf{(b) Unseen resolution}\par Relative $[0,1000]$\par}\smallskip
\includegraphics[width=\linewidth,height=3.2cm,keepaspectratio]{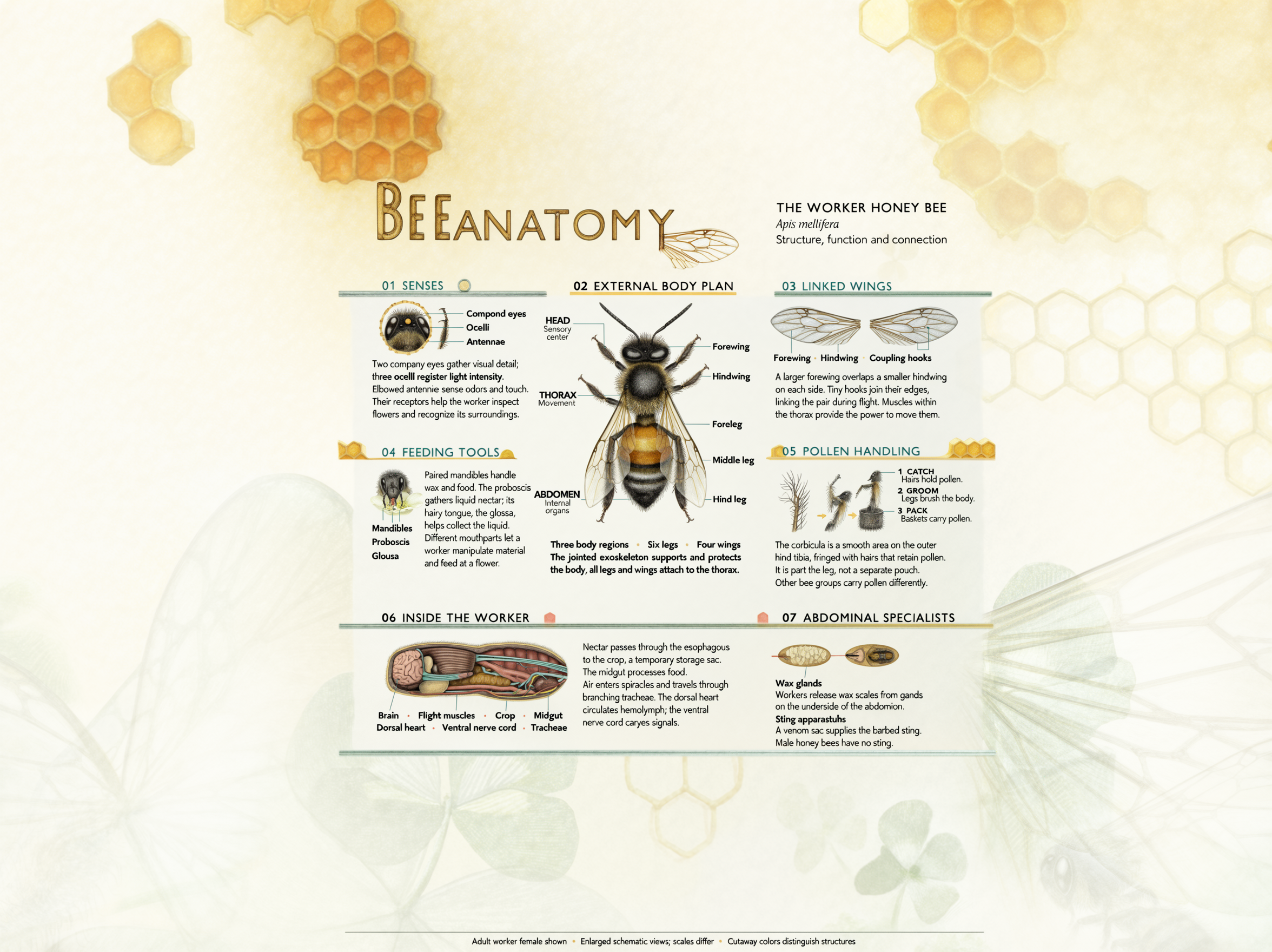}
\end{minipage}\hfill
\begin{minipage}[t]{0.32\linewidth}
\centering
{\small\textbf{(c) Unseen resolution}\par Zero-centered absolute\par}\smallskip
\includegraphics[width=\linewidth,height=3.2cm,keepaspectratio]{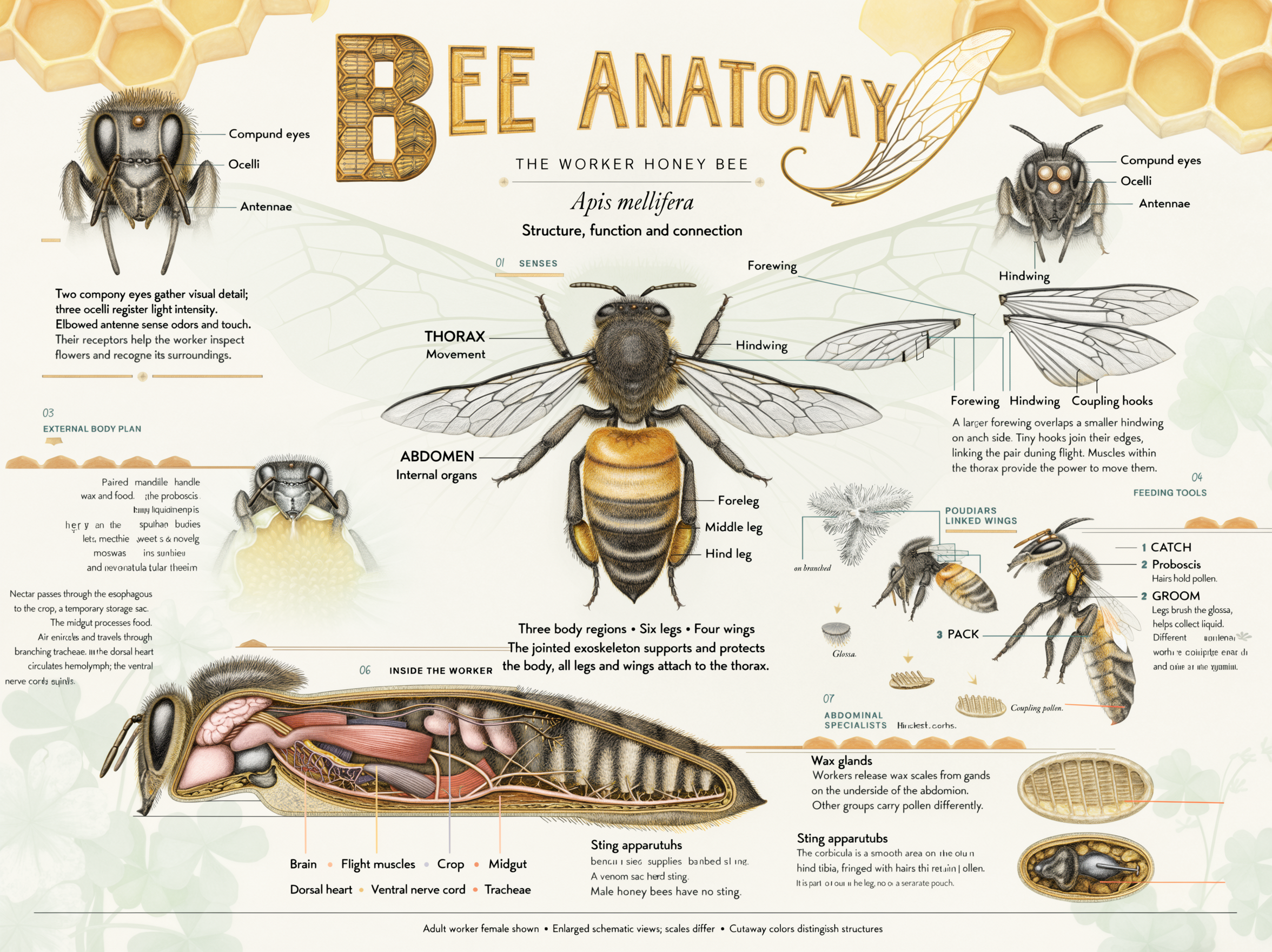}
\end{minipage}
\caption{\textbf{Generalization to a resolution unseen during training.} (a) Lower-resolution comparison. (b) Relative coordinates normalized to $[0,1000]$ concentrate the content into a smaller central region at the unseen resolution. (c) Zero-centered absolute coordinates better preserve the overall spatial scale at the same unseen resolution. Images are displayed at a common height for visual comparison.}
\label{fig:bbox_coordinate_comparison}
\end{figure}

Figure~\ref{fig:bbox_coordinate_comparison} compares a lower-resolution bee-anatomy infographic with two outputs generated at a higher resolution unseen during training, using different bounding-box coordinate representations. Zero-centered absolute coordinates follow the center-relative pixel convention described in Appendix~\ref{app:conditioning}; relative coordinates are normalized to $[0,1000]$.

At the unseen resolution, normalized coordinates in panel (b) concentrate the infographic in a smaller central region, leaving wider margins. Center-relative pixel coordinates in panel (c) distribute the content across the canvas and more closely preserve the spatial scale of the lower-resolution design. This example shows how the coordinate representation affects layout transfer to a larger canvas.

\section{Related Work}
\label{app:related}

\paragraph{Foundation models and pretraining.}
Latent diffusion~\citep{rombach2021highresolution} and diffusion transformers~\citep{peebles2022dit} underpin modern image generation. SDXL~\citep{podell2024sdxl}, Stable Diffusion 3~\citep{esser2024rectified}, FLUX~\citep{flux-2-2025}, Qwen-Image~\citep{wu2025qwen}, Z-Image~\citep{cai2025z}, and ERNIE-Image~\citep{liu2026ernie} advance general-purpose synthesis and instruction following. Qwen-Image-2.1~\citep{qwenimage21} further unifies image generation and editing with a compact 7B visual generator, improved typography, and native support for transparent images. SenseNova-U1.5~\citep{diao2026sensenova15} pursues native unified visual understanding and generation through an encoder-free and VAE-free architecture, with advances in bilingual text rendering, infographic generation, and complex composition. Training recipes are also central: PixArt-$\alpha$~\citep{chen2023pixart} emphasizes informative captions and staged training, while Z-Image combines data curation, multi-level captions, and progressive training. Building on this direction, \ours documents the corpus, multi-granularity and structure-aware supervision, and resolution schedule used to pretrain a 5B generator, with a focus on bilingual long-form text and complex graphics.

\paragraph{Visual text generation.}
GlyphControl~\citep{yang2023glyphcontrol} and TextDiffuser~\citep{chen2023textdiffuser} use glyph or character-level guidance to improve text rendering. AnyText~\citep{tuo2023anytext} extends controlled rendering to multiple languages, and AnyText2~\citep{tuo2024anytext2} adds customizable font and color attributes. TextDiffuser-2~\citep{chen2023textdiffuser2} further uses language models for layout planning and line-level text-position encoding, while Qwen-Image~\citep{wu2025qwen} incorporates strong text rendering into a general-purpose model. These advances motivate our study of long passages within dense, multi-element graphics, where character fidelity and overall composition must be evaluated together.

\paragraph{Layout-conditioned generation.}
ControlNet~\citep{zhang2023controlnet} and T2I-Adapter~\citep{mou2023t2iadapter} introduce spatial control through auxiliary conditioning modules; GLIGEN~\citep{li2023gligen}, LayoutDiffusion~\citep{zheng2023layoutdiffusion}, and BoxDiff~\citep{xie2023boxdiff} explore grounding, layout-aware modeling, and training-free box constraints, respectively. FIBO~\citep{gutflaish2025generating} improves compositional control through long structured captions with descriptive spatial fields, while \ours additionally supports explicit bounding-box coordinates. Spatial alignment alone, however, does not establish accurate long-form text rendering. Closely related concurrent work, Ideogram 4~\citep{ideogram-4-2026}, trains on structured JSON captions with text and layout descriptions, supports bounding-box control and LLM-based prompt expansion, and reports text-rendering and layout results. Its public weight release is limited to FP8/NF4 quantized checkpoints, accompanied by inference code but not a complete pretraining recipe. Our work documents corpus construction, supervision design, and the training schedule for a 5B generator targeting bilingual long passages and dense graphic layouts. \llb and \rla assess rendering under extended text lengths and complete-system performance, respectively.

\paragraph{Prompt enhancement and image agents.}
Promptist~\citep{hao2022promptist} learns to adapt user inputs to model-preferred prompts, and Z-Image~\citep{cai2025z} incorporates reasoning-based prompt enhancement. Concurrent Qwen-Image-Agent~\citep{zhang2026qwen} constructs generation context through planning, reasoning, search, memory, and feedback, including layout-aware content planning. Its training-free framework operates over existing image generators. \ours combines agent-produced layout specifications with a newly pretrained backbone that learns from text-rich data and explicit text-and-structure supervision. Our distinction concerns training the renderer and connecting it to structured layout specifications; prompt-strategy and cross-backbone comparisons examine how these components affect the outputs.

\paragraph{Evaluation of text-rich graphics.}
GenEval~\citep{ghosh2023geneval} and T2I-CompBench~\citep{huang2023t2icompbench} assess compositional alignment. MARIO-Eval~\citep{chen2023textdiffuser}, AnyText-benchmark~\citep{tuo2023anytext}, and LongTextBench~\citep{geng2025xomni} target visual text, while 7Bench~\citep{izzo20257bench} evaluates layout-guided generation and IA-Bench~\citep{zhang2026qwen} tests image-agent capabilities. Our \llb extends bilingual text rendering to longer passages, and \rla evaluates content, layout, and text quality jointly for complete systems responding to realistic graphic-design requests.

\end{document}